%% file: main.tex
\documentclass{article}
\PassOptionsToPackage{square,numbers,sort&compress}{natbib}

\usepackage[final]{neurips_2024}
\usepackage[ruled,vlined]{algorithm2e}

\SetCommentSty{mycommfont}

\usepackage[utf8]{inputenc} 
\usepackage[T1]{fontenc}    
\usepackage[pagebackref=true,breaklinks=true,colorlinks,bookmarks=false]{hyperref}       
\usepackage{url}            
\usepackage{booktabs}       
\usepackage{amsfonts}       
\usepackage{nicefrac}       
\usepackage{microtype}      
\usepackage{xcolor}         

\input{math_commands}

\usepackage{tikz}
\usetikzlibrary{bayesnet}
\usetikzlibrary{arrows}
\usetikzlibrary{positioning}
\usepackage{multirow}
\usepackage{xcolor}
\usepackage{graphicx}
\usepackage{subcaption}
\usepackage{wrapfig}

\title{An Information Theoretic Perspective on \\Conformal Prediction}

\author{%
    Alvaro H.C. Correia \quad
    Fabio Valerio Massoli \quad
    Christos Louizos$^\dagger$ \quad
    Arash Behboodi$^\dagger$ \\
  Qualcomm AI Research\thanks{Qualcomm AI Research, Qualcomm Technologies Netherlands B.V (Qualcomm AI Research is an initiative of Qualcomm Technologies, Inc.). $^\dagger$Equal contribution.} \\
  Amsterdam, The Netherlands \\
  \texttt{\{acorreia, fmassoli, clouizos, behboodi\}@qti.qualcomm.com} \\
}

\begin{document}

\maketitle
\vspace{-0.25cm}

\begin{abstract}
    Conformal Prediction (CP) is a distribution-free uncertainty estimation framework that constructs prediction sets guaranteed to contain the true answer with a user-specified probability. Intuitively, the size of the prediction set encodes a general notion of uncertainty, with larger sets associated with higher degrees of uncertainty.
    In this work, we leverage information theory to connect conformal prediction to other notions of uncertainty.
    More precisely, we prove three different ways to upper bound the intrinsic uncertainty, as described by the conditional entropy of the target variable given the inputs, by combining CP with information theoretical inequalities.
    Moreover, we demonstrate two direct and useful applications of such connection between conformal prediction and information theory: 
    (i) more principled and effective \emph{conformal training} objectives that generalize previous approaches and enable end-to-end training of machine learning models from scratch, and (ii) a natural mechanism to incorporate side information into conformal prediction.
    We empirically validate both applications in centralized and federated learning settings, showing our theoretical results translate to lower \emph{inefficiency} (average prediction set size) for popular CP methods.
\end{abstract}

\section{Introduction}
\input{sections/introduction}

\section{Background} \label{sec:background}
\input{sections/background}

\input{sections/list_decoding}

\section{Conformal Training} \label{sec:confromal_training}
\input{sections/conformal_training}

\section{Side Information} \label{sec:side_information}
\input{sections/side_information}

\section{Related Work} \label{sec:related_work}
\input{sections/related_works}

\section{Experiments} \label{sec:experiments}
\input{sections/experiments}

\section{Conclusion} \label{sec:conclusion}
\input{sections/conclusion}

\bibliography{bibliography.bib}
\bibliographystyle{icml2024}


\appendix
\clearpage
\input{appendix/broader_impact_limitations}
\input{appendix/conformal_background}
\input{appendix/info_theory_background}

\input{appendix/main_results}
\input{appendix/list_decoding}

\input{appendix/conftr}
\input{appendix/experiments}

\input{appendix/neurips_checklist}


\end{document}

%% file: math_commands.tex
\usepackage{amsmath,amsfonts,bm}

\def\eqref#1{eq.~\ref{#1}}

\def\1{\bm{1}}

\def\mZ{{\bm{Z}}}

\DeclareMathAlphabet{\mathsfit}{\encodingdefault}{\sfdefault}{m}{sl}
\SetMathAlphabet{\mathsfit}{bold}{\encodingdefault}{\sfdefault}{bx}{n}

\def\gB{{\mathcal{B}}}
\def\gC{{\mathcal{C}}}
\def\gD{{\mathcal{D}}}

\def\gL{{\mathcal{L}}}

\def\gU{{\mathcal{U}}}

\def\gX{{\mathcal{X}}}
\def\gY{{\mathcal{Y}}}

\def\sR{{\mathbb{R}}}

\usepackage{amsthm}
\usepackage{amssymb}
\newtheorem{theorem}{Theorem}[section]

\newtheorem{lemma}[theorem]{Lemma}
\theoremstyle{definition}
\newtheorem{definition}[theorem]{Definition}

\newtheorem{proposition}[theorem]{Proposition}
\newtheorem{corollary}{Corollary}[section]

\theoremstyle{remark}
\newtheorem{remark}[theorem]{Remark}
\newtheoremstyle{named}{}{}{\itshape}{}{\bfseries}{.}{.5em}{\thmnote{#3}#1}
\theoremstyle{named}

\newtheorem*{rep@theorem}{\rep@title}
\newcommand{\newreptheorem}[2]{%
\newenvironment{rep#1}[1]{%
 \def\rep@title{#2 \ref{##1}}%
 \begin{rep@theorem}}%
 {\end{rep@theorem}}}

\newreptheorem{theorem}{Theorem}
\newreptheorem{proposition}{Proposition}
\newreptheorem{corollary}{Corollary}

\def\P{\mathbb{P}}

\newcommand*{\bracket}[1]{\left[#1\right]}

\newcommand*{\paran}[1]{\left(#1\right)}
\newcommand*{\prob}[1]{\mathbb{P}}

\def\defeq{:=}

\newcommand{\E}{\mathbb{E}}

\newcommand{\R}{\mathbb{R}}

\usepackage{centernot}

\newcommand{\quant}{\text{Quantile}}

\newcommand{\supp}{\mathrm{supp}}

%% file: sections/introduction.tex
Machine learning (ML) models have rapidly grown in popularity and reach, having now found use in many safety-critical domains like healthcare \citep{ahsan2022machine} and autonomous driving \citep{kuutti2020survey}. In these areas, predictions must be accompanied by reliable measures of uncertainty to ensure safe decision-making. However, most ML models are designed and trained to produce only point estimates, which capture only crude notions of uncertainty with no statistical guarantees. Conformal prediction (CP) \citep{vovk2005algorithmic}, in particular its split variant (SCP) \citep{papadopoulos2002inductive}, has recently gained in popularity as a principled and scalable solution to equip \emph{any}, potentially black-box, model with proper uncertainty estimates in the form of prediction sets; in loose terms, larger sets are associated with higher degrees of uncertainty.

In this work, we take a closer look at conformal prediction through the lens of information theory (IT), establishing a connection between conformal prediction and the underlying intrinsic uncertainty of the data-generating process, as captured by the conditional entropy $H(Y|X)= -E_{P_{XY}}[\log P_{Y |X}]$ of the target variable $Y$ given the inputs $X$. We prove conformal prediction can be used to bound $H(Y|X)$ from above in three different ways: one derived from the data processing inequality, which we dub \emph{DPI bound}, and two coming from a variation of Fano's inequality \citep{fano_transmission_1949}, a model agnostic one, which we call (simple) \emph{Fano bound}, and another informed by the predictive model itself, to which we refer as \emph{model-based Fano bound}. To the best of our knowledge, these bounds represent the first bridge connecting information theory and conformal prediction, which we hope will bring new tools to both fields. We already present two such tools in this paper: (i) we show our upper bounds serve as principled training objectives to learn classifiers that are more amenable to SCP, and (ii) we advance a systematic way to incorporate side information into the construction of prediction sets. In a number of classification tasks, we empirically validate that both these applications of our theoretical results lead to better predictive efficiency, i.e., narrower and, consequently, more informative prediction sets.

The rest of the paper is organized as follows. In Section~\ref{sec:background}, we first introduce the necessary background to guide the reader through our main theoretical results. We introduce our three new upper bounds to the intrinsic uncertainty in Section~\ref{sec:main}, and their applications to conformal training and side information in Sections~\ref{sec:confromal_training} and~\ref{sec:side_information}, respectively. Thereafter, we explore the related work in CP and IT in Section~\ref{sec:related_work}, present and analyze our experimental results in Section~\ref{sec:experiments}, and finally conclude in Section~\ref{sec:conclusion}.

%% file: sections/background.tex
In this section, we present the needed background on conformal prediction and list decoding \citep{elias_list_1957,wozencraft_list_1958,guruswami2004list}, an area of information theory that, as we show, is closely related to CP and especially useful in deriving our main results.
We start with the necessary notation. As usual, we denote random variables in uppercase letters and their realization in lowercase,~e.g., $X=x$. We reserve calligraphic letters, e.g. $\gX$, for sets and use $P_{X},Q_X,\dots$ to denote probability measures on the space $\gX$. To simplify the notation, we use $P,Q,\dots$ when the underlying space is clear. For example, given a probability measure $Q_{XY}$, the probability of the event $\{(X,Y): Y\in \gC(X) \}$ is denoted as $Q(Y\in \gC(X))$. 

\subsection{Conformal Prediction}
Conformal prediction (CP) is a theoretically grounded framework that provides \emph{prediction sets} with finite-sample guarantees under minimal distribution-free assumptions. Concretely, given a set of $n$ data points $(X_i, Y_i) \in \gX \times \gY, i=1, \ldots, n$ drawn from some (unknown) joint distribution $P_{XY}$, CP allows us to construct sets $\gC(X) \in \gY$, such that for a new data point from the same distribution $(X_{test}, Y_{test})$ we have the following guarantee for a target error rate $\alpha \in (0, 1)$
\begin{equation} \label{eq:cp}
    \P(Y_{test}\in \gC(X_{test}))\geq 1-\alpha,
\end{equation}
where the probability is over the randomness in the sample $\{(X_i, Y_i)\}_{i=1}^{n}\cup\{(X_{test}, Y_{test})\}$. 
To make this more tangible, the reader can picture $\gC(X)$ as a subset of the possible labels in a classification problem, or as a confidence interval around the point estimate of a regressor in a regression setting.

In this work, we focus on a variant called split conformal prediction (SCP) \citep{papadopoulos2002inductive} that gained popularity in the ML community, since it can leverage \emph{any} pre-trained model $f: \gX \rightarrow \gY$ in the construction of prediction sets. In this setting, the aforesaid $n$ data points constitute a \emph{calibration data set} $\gD_{cal}$, which must be disjoint from the training data set used to fit the predictive model $f$. This separation between training and calibration data is what gives the name \emph{split} to the method.

The first step in SCP is to define a \emph{nonconformity score function} $s_f: \gX \times \gY \rightarrow \sR$, which is itself a function of model $f$ and captures the magnitude of the prediction error at a given data point; the higher the score $s_f(x, y)$, the higher the disagreement between input $x$ and prediction $y$. At calibration time, 
we evaluate the score function at every $(X_i, Y_i) \in \gD_{cal}$ to get a collection of scores $ \{S_i = s_f(X_i, Y_i) \}_{i=1}^n$, 
and at test time, we construct prediction set $\gC(X_{test})$ as
\begin{align} \label{eq:cp_set}
    \gC(X_{test}) = \{y \in \gY : s(X_{test},y) \leq \quant(1-\alpha; \{S_i\}_{i=1}^n \cup \{\infty\})\},
\end{align}
where $\quant(1-\alpha; \{S_i\}_{i=1}^n)$ is the level $1-\alpha$ quantile of the empirical distribution defined by $\{S_i\}_{i=1}^n$. The central result in conformal prediction, which we restate below for completeness, proves that prediction sets thus constructed achieve \emph{marginal valid coverage},~i.e., satisfy (\ref{eq:cp}).

\begin{theorem}[\citep{vovk2005algorithmic, lei2018distribution}] \label{theo:main_cp}
    If $\{(X_i, Y_i)\}_{i}^n$ are i.i.d. (or only exchangeable), then for a new i.i.d. draw $(X_{test}, Y_{test})$, and for any $\alpha \in (0,1)$ and for any score function $s$ such that $\{S_i\}_{i=1}^n$ are almost surely distinct, then $\gC(X_{test})$ as defined above satisfies
        \begin{equation*}
            1 - \alpha \leq \P(Y_{test}\in \gC(X_{test})) \leq 1-\alpha_n, \quad \text{ where} \quad \alpha_n = \alpha - \nicefrac{1}{n+1}.
        \end{equation*}
\end{theorem}
See Appendix~\ref{app:background} for the proof and a more thorough introduction to CP.
It is worth noting that valid coverage is not sufficient; the uninformative set predictor that always outputs $\gC(X_{test})=\gY$ trivially satisfies (\ref{eq:cp}). We would also like our prediction sets to be as narrow as possible, and that is why CP methods are often compared in terms of their (empirical) \emph{inefficiency},~i.e., the average prediction set size $\nicefrac{1}{|D_{test}|}\sum_{x \in \gD_{test}} |\gC(x)|$ for some test data set $\gD_{test}$. This is, in fact, not the only type of inefficiency criterion, but we use it as our main performance metric since it is the most common \citep{vovk2016criteria}.

\begin{wrapfigure}[12]{r}{0.5\textwidth}
    \input{figures/graphical_model}
\end{wrapfigure}

We depict the split conformal prediction procedure in Figure~\ref{fig:graphical-model-cp}, where we include two extra variables that will be useful later in the text: the model prediction  $\hat Y = f(X)$, and the event of valid coverage $E = \{Y \in \gC(X)\}$,~i.e., the event of the prediction set containing the correct class.

\subsection{Conformal Prediction as List Decoding}
In a nutshell, we can see conformal prediction as defining a map from an input $x$ to a set of candidates in the target set $\gY$. It turns out that the conformal prediction framework is equivalent to (variable-size) list decoding, an error-recovery model going back to the works of Elias \citep{elias_list_1957} and Wozencraft \citep{wozencraft_list_1958} in communication theory---we review some of these results in Appendix~\ref{app:list_decoding}. In particular, consider the mapping from the true label $y$ to  the input $x$ as a noisy communication channel $p(x|y)$. The goal of an error-correcting code is then to decode the input $x$ and recover the one true label $y$. List decoding generalizes this idea, allowing the decoder to return a set of outcomes (a list) instead of a pointwise prediction. If the correct \textit{solution} is not part of the set output by the decoder, an error is declared. Although conformal prediction and list decoding were developed for different purposes, namely uncertainty quantification and error correction, it is easy to see that, if we allow for variable-size lists, the list decoding problem for the channel $p(x|y)$ as described above is equivalent to the conformal prediction problem. 

To our knowledge, this link between conformal prediction and information theory (and list decoding in particular) has gone unnoticed in the literature, and in this paper we leverage it in two directions.
First, we apply information-theoretic inequalities for list decoding to upper bound the conditional entropy $H(Y|X)$ of the data-generating process. This leads to new objectives for conformal training (see Section \ref{sec:confromal_training}) and new bounds on the inefficiency of a given model (see Appendices \ref{app:set_size} and~\ref{app:set_size_estimate}). Second, the information-theoretic interpretation of CP gives us an effective and theoretically grounded way of incorporating side-information into CP to improve predictive efficiency (see Section~\ref{sec:side_information}).

%% file: figures/graphical_model.tex
\centering
\tikz{


\node[latent] (x) {$X$};%
\node[latent, right=of x] (y) {$Y$};%
\node[det, above right= 0.5cm of x] (yhat) {$\hat Y$};%
\node[det, above right= 1.75cm of x] (C) {$\gC(X)$};%
\node[latent, left=of C] (Dcal) {$\gD_{cal}$};%
\node[det, right=of y] (E) {$E$};%


\edge {x}{y}
\edge {x}{yhat}
\edge {yhat}{C}
\edge {Dcal}{C}s
\edge {C, y}{E}
 }
\caption{Graphical model of SCP. $\gD_{cal}$ is a calibration set, $\gC(X)$ the prediction set, $\hat Y=f(X)$ the model prediction, and $E$ the event $\{Y \in \gC(X)\}$. Square and round nodes are, respectively, deterministic and stochastic functions of their parents.}%
\label{fig:graphical-model-cp}

%% file: sections/list_decoding.tex
\section{Information Theory Applied to Conformal Prediction} \label{sec:main}
In this section, we develop our main results, which link information theory and conformal prediction. Concretely, we provide three novel upper bounds on the conditional entropy $H(Y|X)$: one coming from the data processing inequality and two derived in a similar way to Fano's inequality.
We defer the proofs to Appendix~\ref{app:proofs} for the sake of conciseness, but in broad strokes, our results come from relating the bounds on the error probability provided by these information-theoretic inequalities (typically in the context of error-correcting codes) to the guarantees provided by CP in Theorem~\ref{theo:main_cp}.

\subsection{Data Processing Inequality for Conformal Prediction}
We start by using the classical data processing inequality (DPI) in the context of conformal prediction. Specifically, we focus on the DPI for $f$-divergences, which we discuss thoroughly in Appendix \ref{app:fdivergence}. In brief, for a convex function $f$ with $f(1)=0$, the $f$-divergence between two probability measures $P$ and $Q$ is defined as (see \citep{sason_f_2016})
\begin{equation*}
    D_f(P||Q)\defeq \E_{Q}\bracket{f\paran{\frac{dP}{dQ}}}.
\end{equation*}
In particular, with $f(x)=x\log x$ we recover the familiar notion of KL-divergence \citep{kullback1951information}. The DPI for $f$-divergences states that for any two probability measures $P_X$ and $Q_X$ defined on a space $\gX$, and any map $W_{Y|X}$, which maps $(P_X,Q_X)$ to $(P_Y,Q_Y)$, we have 
\begin{equation*}
    D_f(P_X||Q_X) \geq D_f\left( P_Y || Q_Y \right).
\end{equation*}
We can apply the DPI for $f$-divergence above in the context of conformal prediction by considering the probability of the event of valid coverage $\{Y \in \gC(X)\}$ under two different probability measures $P$ and $Q$. Taking $P$ as the data-generating distribution $P:=P_XP_{Y|X}$ and constructing $Q:=P_XQ_{Y|X}$ for an arbitrary $Q_{Y|X}$ (e.g., a machine learning model), we get the following proposition. 


\begin{proposition}[DPI Bound] \label{prop:dpi_bound}
Consider any conformal prediction method satisfying the upper and lower bounds of Theorem~\ref{theo:main_cp} for $\alpha\in(0, 0.5)$. For any arbitrary conditional distribution $Q_{Y|X}$, the true conditional distribution $P_{Y|X}$ and the input measure $P_X$, define the following two measures $Q:=P_XQ_{Y|X}$ and $P:=P_XP_{Y|X}$. Then, we have
\begin{align} \label{eq:dpi}
H(Y|X) \leq h_b(\alpha) + \left(1-\alpha \right)\log{Q}(Y \in \gC(X)) + \alpha_n \log {Q}(Y \notin \gC(X)) - \E_{P}\left[\log Q_{Y|X}\right],
\end{align}
with $\alpha_n = \alpha - \nicefrac{1}{n+1}$ and $h_b(\cdot)$ the binary entropy function $h_b(\alpha)=-\alpha\log(\alpha) - (1-\alpha)\log(1-\alpha).$
\end{proposition}


Note that the entropy term $H(Y|X)$ is computed using the measure $P$. We relegate the proof to Appendix~\ref{app:dpi_proof}. In the bound in (\ref{eq:dpi}), the term $Q(Y\in \gC(X))$ appears inside a log, so an empirical estimate $\hat{Q}(Y \in \gC(X))$ would result in a lower bound and would be biased. We can provide an upper confidence bound on this estimate using the empirical Bernstein inequality~\citep{maurer2009empirical} and use that instead. Based on the empirical Bernstein inequality, with probability $1-\delta$, we have
\begin{align*}
    \Delta_\delta(\mathbf{Z}, n) & := \sqrt{\frac{2V_n(\mathbf{Z})\log(2/\delta)}{n}} + \frac{7\log(2/\delta)}{3 (n - 1)}\\
     Q(Y \in \gC(X)) & \leq \hat{Q}(Y \in \gC(X)) + \Delta_\delta(\mathbf{Z}, n) := \tilde{Q}(Y \in \gC(X)),\\
    Q(Y \notin \gC(X)) & \leq \hat{Q}(Y \notin \gC(X)) + \Delta_\delta(\mathbf{Z}, n) := \tilde{Q}(Y \notin \gC(X)),
\end{align*}
with $V_n(\mZ)$ the empirical variance of $\mZ=(Z_1,\dots,Z_n)$,  $Z_i = Q(y_i \in \gC(x_i))$. Using these bounds, we get the following inequality with probability $1-\delta$:
\begin{align}
    H&(Y|X) \leq h_b(\alpha) + \left(1{-}\alpha\right)\log\tilde{Q}(Y \in \gC(X)) + \alpha_n \log \tilde{Q}(Y \notin \gC(X)) - {\E_{P}\left[\log Q_{Y|X}\right]}.
    \label{eq:dpi_bound}
\end{align}
This upper bound is one of our main results, which we dub the \emph{DPI bound}. Note that for the last expectation, we can use the empirical estimate, as it is an unbiased approximation.


\subsection{Model-Based Fano's Inequality and Variations}
Next, we present an inequality which is a variation of Fano's inequality \citep{fano_transmission_1949}, a classical result that, among other things, is used to prove Shannon's classical theorem on channel capacity. See Appendix \ref{app:dpi_fano} for more details. 
In our context, we can use Fano's inequality to relate the conditional entropy $H(Y|X)$ to the probability of error, i.e., $\P(Y \notin \gC(X))$. From that insight, we obtain Proposition~\ref{prop:mbfano} by modifying the classical proof of Fano's inequality, which can be found in \citep{cover_elements_2006}, and applying the conformal guarantees from Theorem~\ref{theo:main_cp} to bound the probability of error. 

\begin{proposition}[Model-Based Fano Bound] \label{prop:mbfano}
Consider any conformal prediction method satisfying the upper and lower bounds of Theorem~\ref{theo:main_cp} for $\alpha\in(0, 0.5)$. Then, for the true distribution $P$, and for any probability distribution $Q$, we have
\begin{multline} \label{eq:mb_fano_bound}
    H(Y|X)  \leq h_b(\alpha) + \alpha\E_{P_{Y, X, \gD_{cal}|Y\notin\gC(X)}}\left[- \log Q_{Y|X, \gC(X), Y\notin\gC(X)}\right] \\
    + \left(1 - \alpha_n\right)\E_{P_{Y, X, \gD_{cal}|Y\in\gC(X)}}\left[- \log Q_{Y|X, \gC(X), Y\in\gC(X)}\right].
\end{multline}
\end{proposition}

Note that we have one term conditioned on the event of valid coverage, $\{Y\in\gC(X)\}$, and another conditioned on $\{Y\notin\gC(X)\}.$ We provide the proof in Appendix \ref{app:model_based_fano}.
A good choice for $Q$ is the predictive model itself, and that is why we refer to the bound above as \emph{Model-Based (MB) Fano bound}. Another natural choice for $Q$ is the uniform distribution, which gives us the following result.

\begin{corollary}[Simple Fano Bound] \label{cor:simple_fano}
    Consider any conformal prediction method satisfying the upper and lower bounds of Theorem~\ref{theo:main_cp} for $\alpha\in(0, 0.5)$. Then, for the true distribution $P$ we have
    \begin{multline}
        H(Y|X) \leq h_b(\alpha) + \alpha\E_{P_{Y, X, \gD_{cal}|Y\notin\gC(X)}}\left[\log (|\gY| - |C(X)|)\right] \\ + \left(1 - \alpha_n\right)\E_{P_{Y, X, \gD_{cal}|Y\in\gC(X)}}\left[ \log |C(X)|\right]. \label{prop:simple_fano}
    \end{multline}
\end{corollary}
The proof follows directly from Proposition~\ref{prop:mbfano} by replacing $Q$ with the uniform distribution.
We refer to the bound in (\ref{prop:simple_fano}) as (simple) \emph{Fano bound}, since it is model agnostic and can be approximated directly with only empirical estimates of the prediction set size. This last bound explicitly relates the central notion of uncertainty in conformal prediction, the prediction set size, to an information-theoretic concept of uncertainty in $H(Y|X)$.
This reinterpretation of conformal prediction as a form of list decoding introduces various information-theoretic tools, potentially useful for various applications. In Appendix~\ref{app:set_size} we derive some new inequalities for conformal prediction, in particular offering new lower bounds on the inefficiency of the conformal prediction. In the next section, we discuss how these inequalities can be used as conformal training schemes. 

%% file: sections/conformal_training.tex
Although split conformal prediction is applicable to any pretrained ML model as a post-processing step, the overall performance of any CP method (commonly measured by its inefficiency) is highly dependent on the underlying model itself. 
Therefore, previous works have proposed to take CP into account already during model training and directly optimize for low predictive inefficiency \citep{colombo2020training,bellotti2021optimized,stutz2022learning}. We use the term \emph{conformal training} to refer to such approaches in general (see Appendix~\ref{app:conftr} for an overview of the topic). In particular, we focus on ConfTr \citep{stutz2022learning} since it generalizes and outperforms \citep{bellotti2021optimized}.  
In ConfTr \citep{stutz2022learning} each training batch $\gB$ is split into calibration $\gB_{cal}$ and test $\gB_{test}$ halves to simulate the SCP process (see  Section~\ref{sec:background}) for each gradient update of model $f$ and minimize the following \emph{size loss}
\begin{equation} \label{eq:size_loss}
    \log \E[|\gC_f(X)|] \approx \log \left(\nicefrac{1}{|\gB_{test}|}\sum_{x \in \gB_{test}}|\gC_f(x)|\right),
\end{equation}
where $\gC_f(x)$ is constructed using the statistics of the nonconformity scores computed on $\gB_{cal}$.
We use the notation $\gC_f(x)$ to emphasize the dependence of the prediction set on model $f$. 
Still, SCP involves step functions and Stutz et al.~\citep{stutz2022learning} introduce two relaxations to recover a differentiable objective:
(i) the computation of quantiles is relaxed via differentiable sorting operators \citep{cuturi2019differentiable,blondel2020fast,petersen2022monotonic}; (ii) the thresholding operation in the construction of prediction sets in (\ref{eq:cp_set}) is replaced by smooth assignments via the logistic sigmoid. The latter relaxation gives ``soft'' prediction sets $\hat \gC_f(x)$, which contain each of the labels $y \in \gY$ with a certain probability. See Algorithm~\ref{alg:conftr} for a depiction of conformal training.

Our upper bounds on $H(Y|X)$, namely DPI, MB Fano and simple Fano, presented in the previous section can be made differentiable in the same way, and thus can also serve as proper loss functions for conformal training. The motivation for doing so is twofold. 
First, the conditional entropy $H(Y|X)$ captures the underlying uncertainty of the task, or equivalently, the uncertainty under the true labelling distribution $P_{Y|X}$. Thus, by minimizing these upper bounds, we can hope to push the model $f$ closer to the true distribution, which is known to achieve minimal inefficiency \citep{vovk2016criteria}.
Interestingly, the cross-entropy loss also bounds $H(Y|X)$ from above and thus can be motivated as a conformal training objective from the same angle. In that regard, the DPI bound from Proposition~\ref{prop:dpi_bound} is particularly advantageous as it is provably tighter than the cross-entropy (see Appendix~\ref{app:dpi_proof}).

Second, we can connect the simple Fano bound from Corollary~\ref{cor:simple_fano} to the size loss (\ref{eq:size_loss}) from \citep{stutz2022learning}. In Appendix~\ref{app:conftr_fano}, we show that via Jensen's inequality and $\log(|Y| {-} |C(X)|) \leq \log|Y|$ the bound in (\ref{prop:simple_fano}) can be further upper bounded as 
\begin{align} \label{eq:upper_bound_conftr}
    \nonumber &\lambda_{\alpha} := h_b(\alpha) + \alpha \log|\gY| - \left(1 - \alpha_n\right) \log(1 - \alpha),\\
    &H(Y|X) \leq \lambda_{\alpha} + \left(1 - \alpha_n\right) \log \E\left[|C(X)|\right],
\end{align}
Since $\lambda_{\alpha}$ and $(1 - \alpha_n)$ are constants, they do not affect optimization, and minimizing the right hand side in (\ref{eq:upper_bound_conftr}) is equivalent to minimizing the size loss in (\ref{eq:size_loss}).
Therefore, we ground ConfTr as minimizing an upper bound to the true conditional entropy that is looser than the simple Fano bound and likely also looser than the model-based Fano bound for an appropriate choice for $Q.$

\input{appendix/algo_poster}

%% file: appendix/algo_poster.tex
\SetInd{0.75em}{0.5em}
\DontPrintSemicolon
\SetKwComment{Comment}{// }{}
\begin{algorithm}[H]
\caption{Conformal training algorithm.} \label{alg:conftr}
\SetKwInOut{Input}{input}
\SetKwInOut{Output}{output}
\Input{A batch of labeled samples $\gB$, a model $f$, a loss function $\gL$ that can be any of our upper bounds or other conformal training objectives.}
\For{\upshape{each training batch} $\gB$}{ 
    split $\gB$ into $\gB_{cal}$ and $\gB_{test}$\;
    \For{\upshape{each} $(x_i, y_i) \in \gB_{cal}$}{
        compute score $s_i = s_f(x_i, y_i)$\;
    }
    sort scores (in a differentiable manner) obtaining $s_{(1)} < s_{(2)} < \ldots < s_{(|\gB_{cal}|)}$\;
    set $\hat q = s_{(\lceil(|\gB_{cal}|+1)(1-\alpha)\rceil)}$  \Comment*[r]{get $1-\alpha$ quantile estimate using $\gB_{cal}$}
    \For{\upshape{each} $(x_j, y_j) \in \gB_{test}$}{
        \For{\upshape{each} $y \in \gY$}{
            $\hat \gC_f(x_j, y) \gets \sigma(\hat q - s_f(x_j, y))$ \Comment*[r]{Construct soft prediction set}
        }
    }
    \Comment*[l]{The DPI and model-based Fano bounds also require class probabilities under $Q$, which in this case are given by the model $f(x_j)$}
    compute loss according to $\gL$ on $\gB_{test}$ using $y_j, \hat \gC_f(x_j)$ and if needed $f(x_j)$ \;
    update $f$ via gradient descent to minimize loss
}
\end{algorithm}

%% file: sections/side_information.tex
With the information theoretical interpretation of conformal prediction, we can translate various intuitions from information theory, for example, about different types of channels or network information theory, to conformal prediction. In this section, we consider the notion of side information. 

Let $X$ be the input covariates, $Y$ be the target variable, $Z$ be some side information about the task, and let $Q_{Y|X}$ be the model which we use to perform conformal prediction. As we can relate CP with $Q_{Y|X}$ to an upper bound on the conditional entropy $H(Y|X)$, we would like to do the same for the case of the conditional entropy when side information is available, i.e., $H(Y|X, Z)$. Since we know that the conditional entropy directly affects the expected set size, i.e., the inefficiency of CP, and given that $H(Y|X) \geq H(Y|X, Z)$ we can expect that with the additional side information the inefficiency of CP will decrease. 
We can take side information into account by defining conformity scores as a function of $Q_{Y|X, Z}$ instead of $Q_{Y|X}$. A simple way to do that would be via the Bayes rule
\begin{align} 
    Q_{Y|X, Z} = \frac{Q_{Y|X} Q_{Z|X, Y}}{\sum_Y Q_{Y|X} Q_{Z|X, Y}},\label{eq:model_with_si}
\end{align}
where $Q_{Z|X, Y}$ is an auxiliary model that predicts the side information given the input and target variables. Such a model could be learned separately from the main model $Q_{Y|X}$ given access to a dataset $\gD_{side} = \{(x_i, y_i, z_i)\}$. 
Therefore, we can now calibrate with CP by taking into account the side information and then, at test time, given access to the input and side information, we can use the appropriate probabilities $Q_{Y|X, Z}$ to construct the prediction sets. 
Intuitively, the prediction sets from such a procedure should be smaller than the prediction sets obtained from using $Q_{Y|X}$ directly. 
It should be noted that, in the case of side information not being available, we can marginalize $Q_{Y, Z|X}$ over $Z$, which, by construction, falls back to the original model $Q_{Y|X}$. If the availability pattern of $Z$ is consistent between the calibration and test sets, the conformal prediction guarantees still hold by defining the model as
\begin{equation} \label{eq:missing_side}
    f = \begin{cases}
      Q_{Y|X, Z} & \text{if $Z$ is observed}\\
      Q_{Y|X} & \text{otherwise.}\\
    \end{cases} 
\end{equation}

With this addition to the split conformal prediction tool set, if new (side) information is made available at test time, we can properly incorporate it into the CP pipeline without having to retrain the main classifier. One needs only train an auxiliary classifier $Q_{Z|X, Y}$, which might be much simpler than $Q_{Y|X}$ (in our experiments, $Q_{Z|X, Y}$ is given by a single linear layer) and require much less data.
One notable example of side information arises naturally in the distributed setting, which we discuss next. 

\subsection{The Distributed Learning Setting}
Consider the case where we have a dataset that is distributed among a set of $m$ devices and want to run conformal training to get a global model $Q_{Y|X}$ trained on all the data. Further, assume it is hard to gather the data at a central location (e.g., due to privacy reasons), and thus we have to work with the data staying locally on each device. An example of this would be federated learning (or FL)~\citep{mcmahan2017communication}. In this case, if $Z \in \{1, \ldots, m\}$ identifies the device, the entropy $H(Y|X)$ can be expressed as
\begin{align*}
     H(Y|X) = H(Y|X, Z) + I(Y;Z|X)
        = \E_{P_Z}\left[H(Y|X, Z=z)\right] + I(Y;Z | X),
\end{align*}
which decomposes into a weighted average of local entropy functions $H(Y|X, Z=z)$. We can now use any of our proposed bounds for each of the conditional entropies $H(Y|X, Z=z)$ by calibrating with CP independently on each device, ending up with
\begin{align*}
    H(Y|X) \leq \E_{P_{Z}}\left[H_{ub}(Y|X, Z=z)\right] + I(Y;Z|X),
\end{align*}
where $H_{ub}(Y|X, Z=z)$ corresponds to an upper bound of the conditional entropy $H(Y|X, Z=z)$. Furthermore, for the mutual information term we have that 
\begin{align*}
    I(Y; Z| X) = \E_{P_{Z, X, Y}}\left[\log \frac{P_{Z|X, Y}}{P_{Z|X}}\right] \leq \E_{P_{Z, X}}\left[- \log P_{Z|X}\right] \leq \E_{P_{Z, X}}\left[- \log Q_{Z|X}\right]
\end{align*}
where the first inequality is due to $Z$ being discrete and having non-negative entropy and the second is due to Gibbs inequality with $Q_{Z|X}$ being an auxiliary model trained to predict the user id $Z=z$ given input $X=x$. A similar upper bound has been considered before in a federated setting \citep{fedsimclr24}. With this upper bound we get
\begin{align} \label{eq:fed_bounds}
    H(Y|X) \leq \E_{P_{Z}}\Big[H_{ub}(Y|X, Z=z) -\E_{P_{X|Z=z}}\left[\log Q_{Z=z|X}\right]\Big].
\end{align}
This gives us an upper bound on the entropy of the entire population that decomposes into a sum of local functions, one for each client, only requiring local information. Thus, we can easily carry out conformal training for $Q_{Y|X}$ by minimizing this upper bound in the federated setting with, e.g., federated averaging~\citep{mcmahan2017communication}. 
At test time, we can take the device ID $z$ as side information. To this end, we can either train a model $Q_{Z|X, Y}$ in parallel and use~(\ref{eq:model_with_si}) with the global model $Q_{Y|X}$ at test time to get $Q_{Y|X, Z}$, or we can obtain $Q_{Y|X, Z}$ by fine-tuning the global model $Q_{Y|X}$ with local data.

%% file: sections/related_works.tex
Conformal prediction, a powerful framework for uncertainty quantification developed by Vovk and collaborators \citep{vovk2005algorithmic,shafer2008tutorial}, has recently witnessed a wide adoption in many fields, e.g., healthcare \citep{papadopoulos2009reliable,alnemer2016conformal,lu2022fair,lu2022improving} and finance \citep{wisniewski2020application,bastos2024conformal}. The marriage of conformal prediction and machine learning has been especially fruitful. Since the seminal work by \citet{vovk2005algorithmic}, many extensions and applications have been proposed, covering topics such as survival analysis~\citep{candes_conformalized_2021}, treatment effect evaluation~\citep{lei_conformal_2021}, classification~\citep{gupta_distribution-free_2020,angelopoulos_uncertainty_2020} and  regression~\citep{romano2019conformalized} settings, risk control~\citep{angelopoulos2021learn,bates_distribution-free_2021}, and covariate shift~\citep{tibshirani_conformal_2019}.

To our knowledge, our work represents the first attempt to bridge conformal prediction and information theory. Among other things, this allows us to build on the conformal training ideas of \citet{bellotti2021optimized} and \citet{stutz2022learning}, deriving principled learning objectives that generalize their approaches, dispense with some of their hyperparameters and result in more efficient prediction sets. 
Further, we empirically show that our conformal training objectives provide a strong enough learning signal to train complex architectures from scratch, with strong results on ResNet-34 and ResNet-50 \citep{he2016deep} fitted on CIFAR10 and CIFAR100, respectively. In contrast, the previous state-of-the-art approach, ConfTr, struggles in those settings (see experiments in Section~\ref{sec:experiments}) and required pretrained models for consistent results \citep{stutz2022learning}. Further, our information-theoretic interpretation of CP provides a new simple and effective mechanism to leverage side information in split conformal prediction. We are unaware of any other approaches to treat side information within the conformal prediction framework in the literature.

On the information theory side, the notion of $f$-divergence and related inequalities have appeared in many different works. The use of $f$-divergence goes back to works of Ali and Silvey, Csisz\'{a}r, and Morimoto in the 60s, as in, for instance, \citep{ali_general_1966,csiszar_topological_1967,morimoto_markov_1963}. A key $f$-divergence inequality is the data processing inequality---see \citep{sason_f_2016,sason_data-processing_2019} for an extensive survey. It provides a unified way of obtaining many classical and new results, including Fano's inequality \citep{fano_transmission_1949}. The tightness of the data processing inequalities is discussed in terms of Bregman's divergence in \citep{collet_exact_2019, liese_divergences_2006} and in terms of $\chi^2$-divergence in \citep{sason_data-processing_2019}. List decoding, which is closely connected to CP, was introduced in the context of communication design by Elias \citep{elias_list_1957} and Wozencraft \citep{wozencraft_list_1958}. A generalization of Fano's inequality to list decoding was given in \citep{csiszar_information_2011} in the context of multi-user information theory, see also the general Fano inequality for list decoding presented in \citep{raginsky_concentration_2013}. Variable-size list decoding was discussed in \citep{sason_data-processing_2019} using ideas first introduced in \citep{polyanskiy_channel_2010} and \citep{liu_e__2017}. A selection of relevant inequalities for list decoding can be found in \citep{sason_data-processing_2019}.

%% file: sections/experiments.tex
In this section, we empirically study two applications of our theoretical results, namely conformal prediction with side information and conformal training with our upper bounds on the conditional entropy as optimization objectives. We focus our experiments on classification tasks since this is the most common setting in previous works in conformal training \citep{colombo2020training,bellotti2021optimized,stutz2022learning}. 

\subsection{Conformal Training} \label{sec:exp_conftr}
We test the effectiveness of our upper bounds as objectives for conformal training in five data sets: MNIST \citep{lecun1998gradient}, Fashion-MNIST \citep{xiao2017fashion}, EMNIST \citep{cohen2017emnist}, CIFAR10 and CIFAR100 \citep{krizhevsky2009learning}. In addition to our three upper bounds, we also evaluate the cross-entropy loss (CE, also another upper bound to the entropy), and the two main variants proposed in \citep{stutz2022learning}, namely ConfTr, which minimizes (\ref{eq:size_loss}) and ConfTr$_{\text{class}}$ that optimizes an additional classification loss term (see Appendix~\ref{app:conftr}). We follow a similar optimization procedure and experimental setup to that of \citep{stutz2022learning}, but with the key differences that we learn the classifiers from scratch in all cases (without the need of pretrained CIFAR models), and that we use the larger ``by class'' split of EMNIST. For each data set, we use the default train and test splits but transfer 10\% of the training data to the test data set. We train the classifiers only on the remaining 90\% of the training data and, at test time, run SCP with 10 different calibration/test splits by randomly splitting the enlarged test data set. See Appendix~\ref{app:experiments} for a complete description of the experimental setup, with extra results and details on model architectures and hyperparameter search. 

\input{tables/conftr}

In Table~\ref{tab:central_inef_results}, we report the empirical inefficiency on test data considering two SCP methods, threshold CP with probabilities (or THR) \citep{sadinle2019least} and APS \citep{romano2020classification}---see Appendix~\ref{app:raps} for results with RAPS \citep{angelopoulos_uncertainty_2020}. In all cases, our upper bounds proved effective loss functions to train efficient classifiers end-to-end and from scratch. For the simpler data sets (MNIST, Fashion-MNIST and EMNIST), all conformal training methods achieved similar results, but both ConfTr methods proved less consistent. This is noticeable in the oftentimes sharp difference in performance between THR and APS, since even after fine-tuning the hyperparameters (see Appendix~\ref{app:experiments}) some of the models failed to converge properly. For the remaining and more challenging data sets, both ConfTr variants lagged behind, probably because they do not provide a strong enough signal to train ResNets from scratch (on CIFAR data sets, \citet{stutz2022learning} only used ConfTr to fine tune pretrained models). A similar observation applies to the simple Fano bound (\ref{prop:simple_fano}), where the relaxed prediction set size is the only learning signal. 

In all experiments, we run conformal training with a target coverage rate of 99\%, i.e., $\alpha=0.01$. It is then important to assess whether the performance of the resulting models deteriorates at different coverage rates, ``overfitting'' to the value of $\alpha$ used for training. In Table~\ref{tab:vary_alpha}, we see how inefficiency varies with $\alpha$ at test time for models trained via conformal training with $\alpha=0.01$. In particular, we can contrast their performance against that of models trained via the CE loss, which is agnostic to the desired coverage rate. In all cases, our model-based Fano and DPI bound performs best with the THR and APS methods, respectively, proving conformal training is worthwhile even if the desired coverage rate might vary at test time. Still, as noticed in \citep{stutz2022learning}, there is a drop in performance in comparison to the CE loss for higher values of $\alpha$ at test time. This could be due to some degree of overfitting, but it could also be attributed to the conformal prediction problem becoming easier for lower coverage rates, thus reducing the gap between our bounds and the CE loss.

\input{tables/vary_alpha} 

\subsection{Side Information}
As a first experiment, we consider datasets for which a natural grouping of the labels exists and use the group assignment as side information $Z$. In CIFAR100, there is a disjoint partition of the 100 classes into 20 superclasses, so we define $z$ as the superclass to which example $(x,y)$ belongs. In EMNIST, $z$ indicates whether the example is a digit, uppercase or lowercase letter. We train an auxiliary model $R_{Z|X, Y}$ and, at test time, assume access to side information $z$ to recompute class probabilities $Q_{Y|X, Z=z}$ from the original classifier $Q_{Y|X}$ as in equation (\ref{eq:model_with_si}). 

We report results for two different scenarios in Table~\ref{tab:ineff_side}. The first is the standard SCP setting, where we assess the inefficiency of THR and APS methods, with side information $Z$ observed for 10, 30 and 100\% of the instances. We redefine the classifier $f$ as in (\ref{eq:missing_side}) to account for when $Z$ is missing, but otherwise, the SCP process remains unchanged. The second scenario is Mondrian or group-balanced CP \citep{vovk2005algorithmic}, where one splits $\gD_{cal}$ into groups and runs CP for each of them individually. In this setting, we group the calibration data points according to $Z$ and base the score function on $Q_{Y|X, Z}$. In all cases, taking the side information into account reduced the inefficiency considerably.
\input{tables/side_info}

\subsection{Federated Learning (FL)}
A practically relevant application of side information arises in FL, where we can take the device ID as side information $Z$. In the federated setting, we train two extra heads on top of the main classifier, one computing $Q_{Z|X}$ so that we can optimize the proper upper bound in (\ref{eq:fed_bounds}), and another computing $Q_{Z|X,Y}$ (while detaching gradients to the main classifier so as to not affect the upper bound optimization) that we use to integrate side information into CP using (\ref{eq:model_with_si}). Besides being a practically relevant application of side information to CP, FL also serves as a more challenging test bed for our conformal training methods, which has not been explored in previous work. We ran federated averaging \citep{mcmahan2017communication} with CE, ConfTr, ConfTr$_{\text{class}}$, and our upper bounds as local loss functions. In this setting, we consider CIFAR10, CIFAR100, and EMNIST with 100, 500, and 1K devices, resp. We assign data points to devices imposing a \emph{distribution-based label imbalance} \citep{li2022federated}, i.e., we sample a marginal label distribution for each device from a Dirichlet distribution $\text{Dir}(1.0)$. See Appendix~\ref{app:experiments} for results with $\text{Dir}(0.5)$ and $\text{Dir}(0.1)$. As hyperparameter search in FL is notably challenging and costly \citep{wang2021field}, we keep the same hyperparameters found for the centralized case in Section~\ref{sec:exp_conftr}.

\input{tables/fed_thr}
\input{tables/fed_aps}

After convergence, we ran SCP with the final global model assuming calibration and test data sets at the server, or equivalently that the clients share their scores with the server. This reflects the best inefficiency results we can hope for with the global model, as in practice we might need to resort to privacy-preserving methods that are likely to hurt performance. See Appendix~\ref{app:experiments} for a discussion and extra results on other possible settings. We report inefficiency results for the global model with THR in Table~\ref{tab:fed_results_thr} (see Table~\ref{tab:fed_results_aps} in the appendix for APS results), where we observe similar trends to the centralized experiments in Table~\ref{tab:central_inef_results}. ConfTr methods still perform well on EMNIST but struggle on CIFAR data (with the notable exception on CIFAR100, where ConfTr excelled) while our methods delivered consistent results across all data sets. This probably reflects the sensitivity of both ConfTr objectives to hyperparameters, which makes them hard to use in practice, especially in FL where hyperparameter optimization is difficult. Conversely, our bounds seem more robust to such variations, as the hyperparameters found in the centralized setting seem to translate well to the federated case.

One marked difference between Tables~\ref{tab:central_inef_results} and~\ref{tab:fed_results_thr} is that the simple Fano bound (\ref{prop:simple_fano}), which lagged behind the DPI bound and its model-based counterpart in the centralized setting, achieved the best results on the federated setting. We hypothesize this could be due to overfitting of the local optimization procedures to the individual data distribution of each device, which hurts the convergence of the global model. This effect is exacerbated on CIFAR100, where we have 500 devices, each of which with very few data points.
The simple Fano bound is less vulnerable to such overfitting since it relies on the main classifier to a much lesser degree. Finally, in almost all cases, our bounds outperformed the CE loss, reassuring the potential of conformal training. Moreover, the inclusion of side information reduced inefficiency in all settings, and markedly so in a few instances. This confirms the effectiveness of our side information approach in a complex and practically relevant scenario, like federated learning.

%% file: tables/conftr.tex
\begin{table*}[!t]
\centering
\caption{\textbf{Inefficiency results for conformal training in the centralized setting.} 
We report the mean prediction set size ($\pm$ standard deviation) across 10 different calib./test splits for $\alpha = 0.01$, showing in bold all values within one std.~of the best result. Results for THR and APS correspond to different models trained with different hyperparameters (see Appendix~\ref{app:experiments}). Lower is better.}
\label{tab:central_inef_results}
\resizebox{\textwidth}{!}{  
\begin{tabular}{l@{\hskip 8pt}c@{\hskip 4pt}c@{\hskip 8pt}c@{\hskip 4pt}c@{\hskip 8pt}c@{\hskip 4pt}c@{\hskip 8pt}  c@{\hskip 4pt}c@{\hskip 8pt}  c@{\hskip 4pt}c}
\toprule
 Method & \multicolumn{2}{c}{MNIST} & \multicolumn{2}{c}{F-MNIST} & \multicolumn{2}{c}{EMNIST} & \multicolumn{2}{c}{CIFAR 10} & \multicolumn{2}{c}{CIFAR 100}\\\midrule
 & THR & APS & THR & APS & THR & APS & THR & APS & THR & APS\\\midrule
CE & $\mathbf{2.29_{\pm0.18}}$ & $2.50_{\pm0.08}$ & $2.39_{\pm0.13}$ & $2.41_{\pm0.17}$ & $\mathbf{2.06_{\pm0.11}}$ & $3.40_{\pm0.18}$ & $\mathbf{1.69_{\pm0.11}}$ & $2.34_{\pm0.22}$& $19.70_{\pm2.05}$ & $26.02_{\pm1.31}$\\\midrule
ConfTr & $6.28_{\pm0.71}$ & $\mathbf{2.10_{\pm0.07}}$ & $\mathbf{1.73_{\pm0.06}}$& $\mathbf{1.89_{\pm0.09}}$& $\mathbf{1.99_{\pm0.10}}$ & $\mathbf{2.36_{\pm0.11}}$ & $9.90_{\pm0.02}$ & $9.98_{\pm0.00}$ & $32.80_{\pm2.75}$& $40.58_{\pm1.23}$\\
ConfTr$_{\text{class}}$ & $\mathbf{2.09_{\pm0.11}}$ & $\mathbf{2.13_{\pm 0.13}}$ & $5.11_{\pm0.49}$& $\mathbf{1.79_{\pm0.07}}$ & $\mathbf{2.01_{\pm0.09}}$ & $\mathbf{2.38_{\pm0.11}}$ & $2.16_{\pm0.09}$ & $2.18_{\pm0.06}$ & $66.48_{\pm3.67}$ & $32.91_{\pm1.53}$ \\\midrule
Fano & $\mathbf{2.09_{\pm0.12}}$ & $\mathbf{2.12_{\pm0.08}}$ & $\mathbf{1.70_{\pm0.05}}$& $\mathbf{1.87_{\pm0.05}}$ & $\mathbf{2.10_{\pm0.11}}$ & $2.75_{\pm0.14}$ & $2.05_{\pm0.05}$ & $2.35_{\pm0.10}$ & $40.30_{\pm 1.10}$& $33.80_{\pm0.93}$\\
MB Fano & $\mathbf{2.24_{\pm0.12}}$ & $2.49_{\pm0.19}$ & $\mathbf{1.80_{\pm0.08}}$ & $2.25_{\pm0.14}$& $\mathbf{2.01_{\pm0.11}}$ & $3.67_{\pm0.13}$ & $\mathbf{1.66_{\pm0.09}}$ & $\mathbf{1.89_{\pm0.06}}$ & $\mathbf{14.61_{\pm0.84}}$ & $21.68_{\pm1.44}$ \\
DPI & $\mathbf{2.24_{\pm0.17}}$ & $2.64_{\pm0.07}$ & $\mathbf{1.73_{\pm0.07}}$ & $2.08_{\pm0.06}$ & $\mathbf{1.98_{\pm0.09}}$ & $4.07_{\pm0.23}$ & $\mathbf{1.64_{\pm0.07}}$ & $\mathbf{1.97_{\pm0.08}}$ & $17.55_{\pm1.33}$ & $\mathbf{17.41_{\pm0.62}}$ \\
\bottomrule
\end{tabular}
}
\end{table*}

%% file: tables/vary_alpha.tex
\begin{table}[!h]
    \centering
        \caption{\textbf{Inefficiency results with varying $\alpha$ at test time}. Average prediction set size on CIFAR100 for different $\alpha$ targets at test time, averaged across 10 random calib./test splits. All methods were only optimized for $\alpha{=}0.01$. The models used for THR and APS might not be the same according to the best hyperparameters found in Table~\ref{tab:hyper_cifar100}. Lower is better.}
        \label{tab:vary_alpha}
        \resizebox{0.75\columnwidth}{!}{
            \begin{tabular}{l c@{\hskip 4pt}c@{\hskip 6pt}c@{\hskip 4pt}c@{\hskip 6pt}c@{\hskip 4pt}c}
                \toprule
                Method & \multicolumn{2}{c}{$\alpha = 0.01$} & \multicolumn{2}{c}{$\alpha = 0.05$} & \multicolumn{2}{c}{$\alpha = 0.1$}\\\midrule
                & THR & APS & THR & APS & THR & APS \\\midrule
                CE & $19.70_{\pm2.05}$ & $26.02_{\pm1.31}$ & $6.11_{\pm0.34}$ & $9.19_{\pm0.34}$ & $3.02_{\pm0.10}$ & $4.52_{\pm0.12}$\\
                \midrule
                ConfTr & $32.80_{\pm2.75}$& $40.58_{\pm1.23}$ & $12.25_{\pm0.47}$ & $21.60_{\pm0.78}$ & $7.13_{\pm0.23}$ & $14.58_{\pm0.47}$\\
                ConfTr$_{\text{class}}$ & $66.48_{\pm3.67}$ & $32.91_{\pm1.53}$ & $14.18_{\pm0.60}$ & $16.80_{\pm0.60}$ & $8.90_{\pm0.40}$ & $11.29_{\pm0.42}$\\
                \midrule
                Fano & $40.30_{\pm1.10}$& $33.80_{\pm0.93}$ & $19.43_{\pm0.80}$ & $16.17_{\pm0.49}$ & $11.46_{\pm 0.58}$ & $9.72_{\pm0.25}$\\
                MB Fano & $\mathbf{14.61_{\pm0.84}}$ & $21.68_{\pm1.44}$ & $\mathbf{5.24_{\pm0.13}}$ & $9.25_{\pm0.30}$ & $\mathbf{2.88_{\pm0.05}}$ & $5.51_{\pm 0.14}$\\
                DPI & $17.55_{\pm1.33}$ & $\mathbf{17.41_{\pm0.62}}$ & $6.26_{\pm0.20}$ & $\mathbf{6.98_{\pm0.32}}$ & $3.33_{\pm0.11}$ & $\mathbf{4.08_{\pm0.14}}$\\
                \bottomrule
            \end{tabular}
        }
\end{table}

%% file: tables/side_info.tex
\begin{table}[!hb]
    \fontsize{9pt}{9pt}\selectfont
    \centering
        \caption{\textbf{Inefficiency results with side information.} We report the mean prediction set size ($\pm$ std.) across 10 different calib./test splits for $\alpha = 0.01$. The side information is the superclass assignment for CIFAR100 and whether the class is a digit / uppercase letter / lowercase letter for EMNIST.}
        \label{tab:ineff_side}
            \begin{tabular}{l@{\hskip 8pt}c@{\hskip 4pt}c@{\hskip 4pt}c@{\hskip 8pt}c@{\hskip 4pt}c@{\hskip 4pt}c}
                \toprule
                Method & \multicolumn{3}{c}{CIFAR 100} & \multicolumn{3}{c}{EMNIST}\\\midrule
                & THR & APS & Acc.(\%) & THR & APS & Acc.(\%)\\\midrule
                CP & $19.70_{\pm 2.05}$ & $26.02_{\pm 1.31}$ & $72.22$ & $2.06_{\pm 0.11}$ & $3.37_{\pm 0.15}$ & $85.74$\\
                CP w/ 10\% SI & $18.13_{\pm 2.63}$ & $23.59_{\pm 2.08} $ & $72.84$ & $1.91_{\pm 0.07} $ & $2.18_{\pm 0.09} $ & $86.93$\\ 
                CP w/ 30\% SI & $15.74_{\pm 1.11}$ & $21.63_{\pm 1.45} $ & $74.83$ & $1.69_{\pm 0.05} $ & $1.88_{\pm 0.07} $ & $89.43$\\ 
                CP w/ 100\% SI & $10.28_{\pm 0.86}$ & $15.65_{\pm 1.17}$ & $78.72$& $1.06_{\pm 0.02}$ & $1.07_{\pm 0.02}$ & $97.65$\\ 
                \midrule
                Group CP & $17.59_{\pm 1.89}$ & $21.92_{\pm 1.80}$ & $72.22$ & $2.32_{\pm 0.14}$ & $2.68_{\pm 0.11}$ & $85.74$\\
                Group CP w/ 100\% SI & $9.07_{\pm 0.60}$ & $13.16_{\pm 0.68}$ & $78.72$ & $1.14_{\pm 0.03}$ & $1.16_{\pm 0.04}$ & $97.65$\\ 
                \bottomrule
            \end{tabular}
\end{table}

%% file: tables/fed_thr.tex
\begin{table*}[!t]
\fontsize{9pt}{9pt}\selectfont
\centering
\caption{
\textbf{Inefficiency results for conformal training in the federated setting with THR.} 
We report the mean prediction set size ($\pm$ standard deviation) of the global federated model across 10 different calib./test splits for $\alpha = 0.01$ and using THR. We use $_{+\text{SI}}$ to indicate the inclusion of side information. We show in bold all values within one standard deviation of the best result. Lower is better.}
\label{tab:fed_results_thr}
\begin{tabular}{ l@{\hskip 8pt} c@{\hskip 4pt}c@{\hskip 8pt}  c@{\hskip 4pt}c@{\hskip 8pt} c@{\hskip 4pt}c@{\hskip 8pt} }
\toprule
 Method & \multicolumn{2}{c}{EMNIST} & \multicolumn{2}{c}{CIFAR 10} & \multicolumn{2}{c}{CIFAR 100} \\\midrule
 & THR & THR$_{+\text{SI}}$  & THR & THR$_{+\text{SI}}$ & THR & THR$_{+\text{SI}}$  \\\midrule
CE & $2.91_{\pm 0.02}$  & $2.46_{\pm 0.02}$ & $2.73_{\pm 0.04}$  & $2.30_{\pm 0.06}$ & $55.41_{\pm 1.09}$  & $52.31_{\pm 1.03}$   \\
\midrule
ConfTr & $4.60_{\pm 0.05}$  & $3.30_{\pm 0.02}$  & $10.00_{\pm 0.00}$  & $10.00_{\pm 0.00}$ & $\mathbf{45.60_{\pm 1.30}}$  & $\mathbf{41.18_{\pm 1.16}}$  \\
ConfTr$_{\text{class}}$ & $2.88_{\pm 0.02}$  & $\mathbf{1.98_{\pm 0.02}}$  & $3.53_{\pm 0.09}$  & $3.39_{\pm 0.08}$  & $58.53_{\pm 1.40}$  & $56.03_{\pm 1.29}$  \\
\midrule
Fano & $\mathbf{2.63_{\pm 0.02}}$  & $2.37_{\pm 0.02}$  & $\mathbf{2.39_{\pm 0.07}}$  & $\mathbf{2.07_{\pm 0.07}}$  & $\mathbf{47.91_{\pm 1.20}}$  & $\mathbf{41.19_{\pm 1.02}}$  \\
MB Fano & $2.84_{\pm 0.04}$  & $2.25_{\pm 0.03}$  & $\mathbf{2.52_{\pm 0.08}}$  & $\mathbf{2.04_{\pm 0.07}}$  & $52.94_{\pm 1.40}$  & $46.97_{\pm 1.30}$   \\
DPI & $\mathbf{2.60_{\pm 0.02}}$  & $2.23_{\pm 0.01}$  & $2.76_{\pm 0.07}$  & $2.28_{\pm 0.03}$  & $52.36_{\pm 0.95}$  & $48.64_{\pm 0.70}$   \\
\bottomrule
\end{tabular}
\end{table*}

%% file: tables/fed_aps.tex
\begin{table*}[!h]
\fontsize{9pt}{9pt}\selectfont
\centering
\caption{\textbf{Inefficiency results for conformal training in the federated setting with APS.} 
We report the mean prediction set size ($\pm$ standard deviation) of the global federated model across 10 different calib./test splits for $\alpha = 0.01$ and using APS. We use $_{+\text{SI}}$ to indicate the inclusion of side information. We show in bold all values within one standard deviation of the best result. Lower is better.}
\label{tab:fed_results_aps}
\begin{tabular}{ l@{\hskip 8pt} c@{\hskip 4pt}c@{\hskip 8pt}  c@{\hskip 4pt}c@{\hskip 8pt} c@{\hskip 4pt}c@{\hskip 8pt} }
\toprule
 Method & \multicolumn{2}{c}{EMNIST} & \multicolumn{2}{c}{CIFAR 10} & \multicolumn{2}{c}{CIFAR 100} \\\midrule
 & APS & APS$_{+\text{SI}}$ & APS & APS$_{+\text{SI}}$ & APS & APS$_{+\text{SI}}$ \\\midrule
CE & $3.69_{\pm 0.03}$ & $3.14_{\pm 0.04}$ & $\mathbf{2.83_{\pm 0.07}}$  & $2.43_{\pm 0.06}$ & $64.73_{\pm 0.34}$  & $62.67_{\pm 3.68}$  \\
\midrule
ConfTr  & $6.14_{\pm 0.04}$  & $5.25_{\pm 0.04}$ & $10.00_{\pm 0.00}$  & $10.00_{\pm 0.00}$ & $55.18_{\pm 2.10}$  & $47.58_{\pm 1.48}$ \\
ConfTr$_{\text{class}}$  & $\mathbf{2.65_{\pm 0.02}}$  & $\mathbf{2.42_{\pm 0.02}}$  & $10.00_{\pm 0.00}$  & $10.00_{\pm 0.00}$ & $99.92_{\pm 0.02}$  & $99.91_{\pm 0.01}$ \\
\midrule
Fano  & $3.12_{\pm 0.04}$  & $2.72_{\pm 0.03}$  & $\mathbf{2.73_{\pm 0.07}}$  & $\mathbf{2.39_{\pm 0.06}}$  & $\mathbf{46.95_{\pm 0.67}}$  & $\mathbf{42.75_{\pm 0.91}}$\\
MB Fano  & $4.75_{\pm 0.03}$  &  $\mathbf{2.43_{\pm 0.01}}$  & $\mathbf{2.79_{\pm 0.13}}$  & $\mathbf{2.33_{\pm 0.05}}$  & $50.72_{\pm 1.77}$  & $45.72_{\pm 1.38}$ \\
DPI  & $2.98_{\pm 0.03}$  & $2.58_{\pm 0.02}$  & $\mathbf{2.68_{\pm 0.15}}$  & $\mathbf{2.22_{\pm 0.09}}$   & $51.29_{\pm 1.07}$  & $47.18_{\pm 1.27}$ \\
\bottomrule
\end{tabular}
\end{table*}

%% file: sections/conclusion.tex
In this work, we established a link between notions of uncertainty coming from conformal prediction and information theory (or more precisely variable-size list decoding). 
We proved that one can use split conformal prediction methods to upper bound the conditional entropy of the target variable given the inputs, and that these upper bounds form principled objectives for conformal training. We empirically validated our approach to conformal training, with strong results in both centralized and federated settings.
Furthermore, the information-theoretic perspective also offers a simple yet rigorous approach to incorporate side information into conformal prediction, which we experimentally show leads to better predictive efficiency.
To the best of our knowledge, this is the first attempt at connecting information theory and conformal prediction. Given the limited communication between these two research communities thus far, we expect our work to incite a fruitful exchange of not only ideas but also theory and algorithms between these two research domains. In this paper, we concentrated our exposition and experiments in classification tasks, but we see an extension of our methods to the regression setting, as a particularly promising avenue for future work.

%% file: appendix/broader_impact_limitations.tex
\section{Broader Impact, Limitations and Other Remarks} \label{app:broader_impact}
\paragraph{Broader Impact.} This work explores the connection between conformal prediction and information theory, with the end goal of advancing the state of the art of uncertainty quantification in machine learning. With that in mind, we believe its potential societal consequences are chiefly positive, since our work might contribute to a larger adoption of uncertainty estimates, in a number of safety-critical applications. Notwithstanding, conformal prediction, like any other uncertainty estimation method, should be applied with care and a proper understanding of the provided guarantees so as not to create an illusion of safety when there is none. That is why our work aims to develop new methods and algorithms from first principles, with the hope of providing new techniques that we can study and understand at a deeper level so as to mitigate, or at least foresee, some of their failure modes.

\paragraph{Limitations.} Our work is not without limitations. In practice, estimating the conditional entropy $H(Y|X)$ is a difficult problem. That is part of the reason upper bounds might be useful, but that also means it is hard to evaluate how tight our upper bounds are. This also limits potential applications of our upper bounds like, for instance, estimating the expected prediction set size. Our experiments in Appendix~\ref{app:set_size_estimate} rely on quantization to get a reasonable lower bound on the conditional entropy, and thus lower bound the expected prediction set size. Finally, in terms of our experimental results, we unfortunately have not been able to single out which of our new upper bounds performs best. The general trend we observe is that the DPI and model-based Fano bounds perform better on complex classification tasks, while simple Fano seems to excel in relatively simple tasks or where there is a high risk of overfitting, like in the federated setting. 

\paragraph{Regression Setting.} In this paper, we focus our experiments on the classification setting,  similarly to previous works on conformal training \citep{colombo2020training,bellotti2021optimized,stutz2022learning}. However, in principle, our bounds pose no assumptions on the underlying prediction problem and we see the application of our results to the regression setting as a promising avenue for future work. In practice, only the simple Fano bound in (\ref{prop:simple_fano}) would not be directly applicable to the regression setting, but that is mainly because it assumes a uniform distribution over the output space and we get the, potentially infinite, $|\mathcal Y|$ term. The other two bounds we propose, DPI in (\ref{eq:dpi_bound}) and model-based Fano in (\ref{eq:mb_fano_bound}), do not include the $|\mathcal Y|$ and $|C(X)|$ terms and can be applied to regression problems as is.

\paragraph{Computational Cost.} The computational cost of using our bounds for conformal training is the same as that of previously proposed conformal training \citep{stutz2022learning}. In comparison to regular training, i.e. minimizing the cross-entropy loss, the additional cost is given by the (differentiable) sorting operation of the scores, which in our case was performed with diffsort \cite{petersen2022monotonic} with bitonic networks, which has complexity $\mathcal O(b^2 \log b^2)$ for $b$ the batch size. Since the batch size is typically small, the additional computation cost is only marginal in practice.

%% file: appendix/conformal_background.tex
\section{Background on Conformal Prediction} \label{app:background}
In this section, we outline a brief introduction to conformal prediction, providing the reader with the necessary background to follow our main results. Readers already familiar with conformal prediction can safely skip this section. We start by reviewing the definitions of quantiles and exchangeability, which are central to the main results in conformal prediction.
\subsection{Exchangeability, Ranks and Quantiles} 

The main assumption in conformal prediction is that the data points used for calibration and testing are exchangeable. Next, we define the concept of exchangeability and discuss how it leads to the main results in conformal prediction via properties of ranks of exchangeable random variables. Our exposition is markedly brief, and we refer the reader to \cite{kuchibhotla2020exchangeability} for a more thorough discussion on exchangeability and its importance in conformal prediction. 

Formally, the concept of exchangeability can be defined as follows.

\begin{definition}[Exchangeable Random Variables]
Random variables $X_1, \ldots, X_n$ are said to be exchangeable if for any permutation $\pi: \{1, \ldots, n\} \rightarrow \{1, \ldots, n\}$, the sequences $(X_1, \ldots, X_n)$ and $(X_{\pi(1)}, \ldots, X_{\pi(n)})$ have the same joint probability distribution.
\end{definition}

Note that exchangeability is a weaker assumption than the i.i.d.~(independent and identically distributed) assumption commonly relied upon in machine learning. More precisely, exchangeable random variables must be identically distributed but not necessarily independent \cite{kuchibhotla2020exchangeability}. Naturally, i.i.d.~random variables are also exchangeable.

One relevant consequence of exchangeability that is central to conformal prediction is that the ranks of exchangeable random variables are uniformly distributed. We define ranks and this property more formally in Definition~\ref{def:rank} and Lemma~\ref{lemma:uniform}, respectively.

\begin{definition}[Rank]\label{def:rank}
    For a set of $n$ elements $\gX = \{x_1, \ldots, x_n\}$,  the rank of any one element $x_i$ in $\gX$ is defined as 
    \[
    \text{rank}(x_i;\gX) = |\{j \in \{1, \ldots, n\} :x_j + \xi U_j \leq x_i + \xi U_i\}|,
    \]
    for $\xi \geq 0$ and $\gU = \{U_1, \ldots, U_n\}$ a set of i.i.d.~random variables uniformly distributed in $[-1, -1].$
\end{definition}

\begin{remark}
    The addition of i.i.d.~uniform noise serves as a tie-breaking mechanism. Since $\{U_1, \ldots, U_n\}$ are almost surely distinct, $\{x_i + \xi U_i\}_{i=1}^n$ are also distinct with probability one. This is necessary to render the rank independent of the distribution of $X_i$, which is key to ensure the distribution-free quality of conformal prediction.
\end{remark}

\begin{lemma}\label{lemma:uniform}
    If $(X_1, \ldots, X_n)$ are exchangeable random variables, then 
    \[
    \left( \emph{\text{rank}}(X_i; \{X_1, \ldots, X_n\}) \right)_{i=1}^n \sim \emph{\text{Unif}}\left(\{ \pi: \{1, \ldots, n\} \rightarrow \{1, \ldots, n\}\}\right).
    \]
\end{lemma}
In words, Lemma~\ref{lemma:uniform} tell us that the ranking of exchangeable random variables is uniformly distributed among all possible permutations $\pi: \{1, \ldots, n\} \rightarrow \{1, \ldots, n\}.$ That means the probability of observing any one ranking is equal to $\nicefrac{1}{n!}$ and, importantly, independent of the distribution of $X$. The corollary below follows directly from Lemma~\ref{lemma:uniform}.

\begin{corollary}\label{cor:uniform}
    If $(X_1, \ldots, X_n)$ are exchangeable random variables, then 
    \begin{equation*}
        \mathbb{P}\left(\text{rank}(X_i; \{X_1, \ldots, X_n\}) \leq t\right) = \frac{\lfloor t \rfloor}{n},
    \end{equation*}
         for $t \in [0, n]$ and $\lfloor t \rfloor$ the smallest integer smaller or equal to $t$. Moreover, we can define a valid p-value as $P:= \text{rank}(X_i; \{X_1, \ldots, X_n\})/n$ since 
    \begin{equation*}
        \mathbb{P}\left(P \leq \alpha \right) \leq \alpha \quad \text{for all } \alpha \in [0, 1].
    \end{equation*}
\end{corollary}
\begin{proof}
    \begin{align*}
        \mathbb{P}(\text{rank}(X_i; \{X_1, \ldots, X_n\}) \leq t) &= \mathbb{P}(\text{rank}(X_i; \{X_1, \ldots, X_n\}) \leq \lfloor t \rfloor) \\
        &= \sum_{i=1}^{\lfloor t \rfloor} \mathbb{P}(\text{rank}(X_i; \{X_1, \ldots, X_n\}) = i) \\
        &= \sum_{i=1}^{\lfloor t \rfloor} \frac{(n-1)!}{n!} = \frac{\lfloor t \rfloor}{n},
    \end{align*}
where the first equality follows because $\text{rank}(.)$ returns an integer, and the third equality follows directly from Lemma~\ref{lemma:uniform}: each permutation of $\left(\text{rank}(X_i; \{X_1, \ldots, X_n\}) \right)_{i=1}^n$ has the same probability $\nicefrac{1}{n!}$, and there are $(n-1)!$ configurations where $\text{rank}(X_i; \{X_1, \ldots, X_n\}) = i$ since $X_i$ is fixed at rank $i$, and we can permute the other $(n-1)$ variables.
\end{proof}

As we shall see, Corollary~\ref{cor:uniform} is central to the main result in conformal prediction, but before proving that result, we should first define the concept of quantiles.

\begin{definition}[Quantile]
For $Z$ a random variable with probability distribution $F$, the level $\beta$ quantile of distribution $F$ is defined as
\[
\quant(\beta;F)=\inf\{z: \P\{Z\leq z\}\geq \beta\}.
\]
Similarly, for a sample $\{z_i\}_{i=1}^n$ and $\delta_{z_i}$ a point mass concentrated at $z_i$,
the quantile of the empirical distribution is given by
\[
\quant\left(\beta;\{z_i\}_{i\in [n]}\right) = \quant\left(\beta; \frac{1}{n}\sum_{i=1}^n\delta_{z_i}\right).
\]
\end{definition}

\subsection{Conformal Prediction} \label{app:background_cp}

Armed with the definitions of quantiles and exchangeability, we are prepared to study conformal prediction, a distribution-free uncertainty quantification framework with the following goal: given a set of data points $\{(X_i, Y_i)\}_{i=1}^n$ sampled from some distribution $P$ on $\gX \times \gY$, to construct a set predictor $\gC: \gX \rightarrow 2^{\gY}$ such that for a new data point $(X_{test}, Y_{test})$ and target error rate $\alpha \in (0, 1)$, we have the guarantee
\[
\P(Y_{test}\in \gC(X_{test}))\geq 1-\alpha,
\]
where the probability is over the randomness of $\{(X_i, Y_i)\}_{i=1}^{n}\cup\{(X_{test}, Y_{test})\}$. Since the probability is taken over both $X_{test}$ and $Y_{test}$, this means that \emph{on average} across all possible values in $\gX$, the correct label $Y_{test}$ is included in the constructed set $\gC(X_{test})$ with probability at least $1-\alpha.$ This property is known as \emph{marginal coverage}. This is in contrast to the stronger \emph{conditional coverage}, which requires
\[
\P(Y_{test}\in \gC(X_{test}) | X_{test})\geq 1-\alpha.
\]
That is, the guarantee holds for each $X_{test}$ individually, with $X_{test}$ fixed and the probability taken over the randomness of $Y_{test}$ only. However, distribution-free conditional coverage is known to admit no non-trivial solution (i.e., besides $\gC(X_{test}) = \gY$) \cite{vovk2012conditional,lei2014distribution}. Therefore, in this paper, whenever we refer to the lower and upper bounds provided by conformal prediction, we mean those provided by Theorem~\ref{theo:main_cp}, which guarantees only marginal coverage.

There are different ways to achieve marginal coverage, but for simplicity we will focus on split conformal prediction or SCP \cite{papadopoulos2002inductive}, since it is easier to grasp than other variants of conformal prediction and is also the main object of study in this paper. We start by assuming a calibration dataset $\gD_{cal}$ which consists of $n$ i.i.d. samples $(X_i,Y_i)$ drawn from an unknown distribution over $\gX\times\gY$. We also assume access to a model $f:\gX\to\hat{\gY}$, where the output space $\hat{\gY}$ can be different from $\gY$. Prediction sets satisfying the guarantee above can be constructed via the following three steps.
\begin{enumerate}
    \item Define a \emph{non-conformity} score function $s: \gX \times \gY \rightarrow \mathbb R,$ which assigns high scores to unusual pairs $(x, y)$. The score function is typically a function of the model $f$ itself.
    \item Compute $S_i = s(X_i, Y_i)$ for all $(X_i, Y_i) \in \gD_{cal}$ and compute $\quant(1-\alpha; \{S_i\}_{i=1}^n \cup \{\infty\})$, the empirical $1-\alpha$ quantile of the scores in $\gD_{cal}$.
    \item Construct prediction sets $\gC(X_{test}) = \big\{y: s(X_{test}, y) \leq \quant(1-\alpha; \{S_i\}_{i=1}^n \cup \{\infty\}) \big\}$
\end{enumerate}

\begin{reptheorem}{theo:main_cp}
    (same result as in the main text (\citet{vovk2005algorithmic, lei2018distribution}) \\
    Let $\{(X_i, Y_i)\}_{i}^n$ be i.i.d. (or only exchangeable) random variables, and $\{S_i\}_{i=1}^n$ be the corresponding set of scores $S_i = s(X_i, Y_i)$ given to each pair $(X_i, Y_i)$ by some score function $s: \gX \times \gY \rightarrow \mathbb R$. Then for a new i.i.d. draw $(X_{test}, Y_{test})$ and any target error rate $\alpha \in (0,1)$, the prediction set constructed as 
        \begin{equation*}
            \gC(X_{test}) = \big\{y: s(X_{test}, y) \leq \emph{\quant}(1-\alpha; \{S_i\}_{i=1}^n \cup \{\infty\}) \big\}
        \end{equation*}
    satisfies the marginal coverage property 
        \begin{equation*}
            \P(Y_{test}\in \gC(X_{test}))\geq 1-\alpha,
        \end{equation*}
    Moreover, if $\{S_i\}_{i=1}^n$ are almost surely distinct, this
    probability is upper bounded by $1 - \alpha + \nicefrac{1}{n + 1}$.        
\end{reptheorem}
\begin{proof} For simplicity, we assume that the set of scores $\{S_i\}_{i=1}^n$ are distinct (or have been made distinct by a suitable random tie-breaking rule). Denote $S_{test} = s(X_{test}, Y_{test})$ and observe that since $s(\cdot)$ is applied element-wise to each pair $(X_i, Y_i)$ it preserves exchangeability, and thus random variables $\{S_i\}_{i=1}^n \cup \{S_{test}\}$ are also exchangeable. Next, we show that the following events are all equivalent
\begin{align*}
    Y_{test} \in \gC(X_{test}) &\stackrel{(i)}{\Longleftrightarrow} S_{test} \leq \quant(1-\alpha; \{S_i\}_{i=1}^n \cup \{\infty\}) \\
    &\stackrel{(ii)}{\Longleftrightarrow} S_{test} \leq \quant(1-\alpha; \{S_i\}_{i=1}^n \cup \{S_{test}\}) \\
    &\stackrel{(iii)}{\Longleftrightarrow} \text{rank}(S_{test}; \{S_i\}_{i=1}^n \cup \{S_{test}\}) \leq \lceil (n+1)(1-\alpha)\rceil.
\end{align*}
\begin{itemize}
    \item[(i)] follows from the construction of the prediction set itself $\gC(X_{test})$ itself.
    \item[(ii)] can be easily verified as follows. 
    If $S_{test} \leq \quant(1-\alpha; \{S_i\}_{i=1}^n \cup \{\infty\}),$ then shifting all values $S_i \geq S_{test}$ to arbitrary values larger than $S_{test}$ will not change the validity of the inequality, since the $1-\alpha$ quantile remains unchanged. In particular, the inequality holds when replacing $\{S_{test}\}$ with $\{\infty\}$ and vice-versa.
    \item[(iii)] follows from the fact that if $S_{test} \leq \quant(1-\alpha; \{S_i\}_{i=1}^n \cup \{S_{test}\})$, then $S_{test}$ is among the $\lceil (n+1)(1-\alpha)\rceil$ smallest values of the set $\{S_i\}_{i=1}^n \cup \{S_{test}\}.$
\end{itemize}
Finally, this is where the crucial exchangeability assumption comes into play, allowing us to apply Corollary~\ref{cor:uniform} to get
\begin{equation*}
    \mathbb P(Y_{test} \in \gC(X_{test})) = \mathbb P(\text{rank}(S_{test}; \{S_i\}_{i=1}^n \cup S_{test}) \leq \lceil (n+1)(1-\alpha)\rceil) 
    = \frac{\lceil (n+1)(1-\alpha)\rceil}{n+1}
\end{equation*}
From there, it is easy to verify that the right hand side is at least $1-\alpha$ and at most $1-\alpha + \nicefrac{1}{n+1}.$
\end{proof}

%% file: appendix/info_theory_background.tex
\section{Background on List Decoding and Basic Information Theoretic Results}
\label{app:info_background}
In this section, we enunciate some classical results from information theory that are instrumental in deriving our main theoretical results, as we show in the detailed proofs in Appendix~\ref{app:proofs}. We also provide a brief introduction to list decoding and demonstrate conformal prediction can be framed as a list decoding problem.

\subsection{Data processing inequalities for f-divergence} \label{app:fdivergence}
We start by presenting the data processing inequality (DPI) for $f$-divergence, which we define below.

\begin{definition} [$f$-Divergence]
Consider two probability measures $P$ and $Q$ and assume that the measure $P$ is absolutely continuous with respect to $Q$, i.e., $P \ll Q$. For a convex function $f:(0,\infty)\to\R$ with $f(1)=0$, the $f$-divergence is defined as:
\begin{equation*}
    D_f(P||Q)\defeq \E_{Q}\bracket{f\paran{\frac{dP}{dQ}} },
\end{equation*}
where $\frac{dP}{dQ}$ is a Radon-Nikodym derivative. In particular, using $f(x)=x\log x $ we recover the familiar notion of KL-divergence
\begin{equation*}
    D_{KL}(P||Q)\defeq \E_{P}\bracket{\log\paran{\frac{dP}{dQ}} }.
\end{equation*}
\end{definition}

We can similarly define conditional $f$-divergence.
\begin{definition} [Conditional $f$-Divergence]
Consider two probability measures $P$ and $Q$ such that $P := P_X P_{Y|X}$ and $Q := P_X Q_{Y|X}$ and that the measure $P$ is absolutely continuous with respect to $Q$. For a convex function $f:(0,\infty)\to\R$ with $f(1)=0$, the conditional $f$-divergence is defined as:
\begin{equation*}
    D_f(P_{Y|X}||Q_{Y|X}|P_X) \defeq \E_{P_X}\bracket{D_f(P_{Y|X=x}||Q_{Y|X=x})}.
\end{equation*}
\end{definition}

Theorem~\ref{theo:dpi} is the classical data processing inequality (DPI), which we restate below for the sake of completeness.
The classical versions of DPI, stated in terms of mutual information, can be found in classical information theoretic text books like \citep{cover_elements_2006}, while the generalization of DPI to $f$-divergences can be found in other works with a comprehensive survey in \citep{polyanskiy2014lecture,sason_f_2016}.

\begin{theorem}[Data Processing Inequality for $f$-divergence] \label{theo:dpi} Consider a conditional distribution $W_{Y|X}$. Suppose that $P_Y$ and $Q_Y$ are two distributions obtained by marginalization of $P_XW_{Y|X}$ and $Q_XW_{Y|X}$ over $X$. For any $f$-divergence, we have
\[
D_f(P_X||Q_X)\geq D_f(P_Y||Q_Y).
\]
\end{theorem}
The proof can be found in standard textbooks in information theory see, for example, Chapter 7, Section 7.2 in \citep{polyanskiy2014lecture} for a derivation of the DPI enunciated in the same form as above. Next, we consider the application of the DPI to conformal prediction.

\begin{theorem}[Data Processing Inequality for Conformal Prediction] \label{theo:dpi_cp}
For any set function $\gC:\gX\to 2^{\gY}$, any $f$-divergence, and all distribution pairs $P,Q$ on $(X,Y)$, we have:
\begin{equation*}
    D_f(P||Q) \geq d_f\left( P(Y\in \gC(x)) || Q(Y\in \gC(x)) \right),
\end{equation*}
where $d_f(p||q)$ is the binary $f$-divergence, namely $d_f(p||q) = q f(p/q)+(1-q)f(1-p/1-q)$.
\end{theorem}
\begin{proof}
    Consider the random variable $E=\mathbf{1}(Y\in \gC(x))$ denoting the event of valid coverage, and the conditional distribution $P_{E|X,Y}=\E_{\gD_{cal}}[\mathbf{1}(Y\in \gC(x))|X,Y]$. Note that $P_{E|X,Y}$ maps distributions $P_{X,Y}$ and $Q_{X,Y}$ to $P_E$ and $Q_E$, resp. From here, the result follows directly from Theorem~\ref{theo:dpi}.
\end{proof}
The construction in Theorem~\ref{theo:dpi_cp} can be further improved as follows.

\begin{theorem}[Conditional Data Processing Inequality for Conformal Prediction] \label{theo:cond_dpi_cp}
For any set function $C:\gX\to 2^{\gY}$, any $f$-divergence, and all \textit{conditional} distribution pairs $P_{Y|X},Q_{Y|X}$, and $P_X$, we have:
\begin{equation*}
    \E_{P_X} D_f(P_{Y|X=x}||Q_{Y|X=x}) \geq \E_{P_X} d_f\left( P_{Y|X}(Y\in \gC(x)|X=x) || Q_{Y|X}(Y\in \gC(x)|X=x)\right),
\end{equation*}
where $d_f(p||q)$ is the binary $f$-divergence, namely $d_f(p||q) = q f(p/q)+(1-q)f(1-p/1-q)$.
\end{theorem}
\begin{proof}
    Consider the conditional distribution $P_{E|X=x,Y}=\E_{\gD_{cal}}[\mathbf{1}(Y\in \gC(x))|X=x,Y]$. We have that
    \begin{align*}
    D_f(P_{Y|X=x}||Q_{Y|X=x}) &= D_f(P_{Y|X=x}P_{E|X=x,Y}||Q_{Y|X=x}P_{E|X=x,Y})\\
    & = D_f(P_{Y,E|X=x}||Q_{Y,E|X=x})
    \end{align*}
    and from the monotonicity property of  f-divergences~\citep{polyanskiy2014lecture} we have that
    \begin{align*}
    D_f(P_{Y|X=x}||Q_{Y|X=x}) &= D_f(P_{Y,E|X=x}||Q_{Y,E|X=x}) \\
    &\geq D_f(P_{E|X=x}||Q_{E|X=x}) \\
    &= d_f(P_{Y|X}(Y \in C(x)|X=x)||Q_{Y|X}(Y\in C(x)|X=x)).
    \end{align*}
    By taking the expectation with respect to $P_X$, we conclude the proof.
\end{proof}

\subsection{List Decoding} \label{app:list_decoding}
List decoding \citep{elias_list_1957,wozencraft_list_1958} is a notion coming from coding theory, a large branch of engineering and mathematics that arises from the application of information theory to the design of reliable communication systems and robust information processing and storage.
In particular, we are interested in channel coding, a field concentrated on the design of so-called error-correcting codes to enable reliable communication over inaccurate or noisy communication channels. 

The general setup studied in channel coding, including list decoding, can be summarized as follows. A message $y \in \gY$ is encoded and transmitted over a noisy channel, and a message $x \in \gX$ is received. The noisy channel is governed by probability density $p(x | y)$ that describes the probability of observing output $x \in \gX$ given input $y \in \gY$. The receiver attempts to \emph{decode} $x$, that is, to guess the originally transmitted message $y$ from the received one, $x$. 
At this point, parallels to machine learning should already have become clear. If the receiver provides a single guess of the transmitted message $y$, we are in a unique-decoding scenario, which is akin to a point prediction in machine learning. Conversely, if the receiver is allowed to guess a set (or a list in the list decoding formalism) of the most likely messages, we have list decoding, which closely resembles conformal prediction. 
Note that in many settings, a list decoding algorithm is constrained to output a list of fixed size. While simple conformal prediction methods for regression settings show the same limitation, the parallel between the two domains is more pertinent when we consider \emph{variable-size} list decoding \citep{sason_data-processing_2019}.

More formally, a list-decoding algorithm can be defined by a set predictor $L: \gX \rightarrow 2^{\gY}$, with maximum output set size $|L(X)| \leq M.$ Naturally, the goal is to design the function $L$ so as to maximize the probability of $Y \in L(X).$ Similarly, for a given input-label pair $(X,Y)$ the goal of conformal prediction is to provide a set that contains $Y$ with a certain pre-determined probability. It is this connection that motivates our bounds on the conditional entropy $H(Y|X).$ 
However, the nonconformity score in conformal prediction is typically a function of the output of a given model $f: \gX \rightarrow \gY$. For instance, $\hat{Y} =f(X)$ could be the logits output by the model, i.e., a vector with the dimension equal to number of classes. 
If we consider $\hat{Y}$ the noisy observation of the ground-truth $Y$, then the problem of determining a set containing $Y$ is again the list decoding problem. Therefore, we can also consider the communication channel $p(\hat{y}|y; x)$ directly, which justifies applying the same upper bounds to $H(Y|\hat Y).$ Note that in this last case, the input data $X$ can also be taken as side information, although it is rarely used directly in building the conformal prediction set. 

This reinterpretation of conformal prediction as list decoding allows us to apply some of the standard results from the list decoding literature to conformal prediction, as we show in the next section.

\subsection{List decoding: Information Theoretic Inequalities}
\textbf{Fano's inequality for variable size list decoding.} The following generalization of Fano's inequality is given in  [\citep{raginsky_concentration_2013}, Appendix 3.E].
\begin{theorem}
Consider a scenario where a decoder upon observing $\hat Y$ provides a nonempty list $L(\hat Y)$ that contains another random variable $Y \in \gY$ with $|\gY|=M$. Define $P_e:=\P(Y \notin L(\hat Y))$. We have:
\begin{equation*}
        H(Y|\hat Y)\leq h_b(P_e) + P_e\log(M) + \E(\log|L(\hat Y)|).
\end{equation*}
\label{thm:fano_list_decoding}
\end{theorem}

\textbf{Optimal list decoding and conformal prediction.} It can be shown that the optimal list decoder consists of selecting $L(\hat y)$ elements of $\gY$ with highest conditional probability $p(y|\hat y)$. That is, consider the sorted posteriors under the true distribution $p(y_1|\hat y)\geq p(y_2|\hat y)\geq \dots \geq p(y_M| \hat y)$ and choose the first $\{y_1,\dots,y_{|L|}\}$ for some list size $|L|$. However, this rule is for fixed-size list decoding and does not determine how to select the coverage set size to guarantee a given coverage. We can modify this rule to obtain a variable-size list decoding with the required coverage. Assuming again the sorted posteriors $p(y_1|\hat y)\geq p(y_2|\hat y)\geq \dots \geq p(y_M| \hat y)$, we can select the set as follows:
\begin{equation*}
    L(\hat y) =\{y_1,\dots,y_{\ell_y}\} \quad \text{ where } \ell_y:=\inf\left\{j:\sum_{i=1}^j p(y_i|\hat y)\geq 1-\alpha\right\}.
\end{equation*}
It is easy to see that the above set is the smallest set given each $y$ and the confidence level $1-\alpha$ (see, for example, \citep{merhav_list_2014}). The same result holds in conformal prediction \citep{vovk2016criteria,romano2020classification}.

\subsection{Fano and Data Processing Inequalities for Conformal Prediction}  \label{app:dpi_fano}

First, from Fano's inequality for list decoding, Theorem \ref{thm:fano_list_decoding}, we get the next result ``out-of-the-box''.
\begin{proposition} \label{prop:fano_list_decoding}
Suppose that $|\gY|=M$. Any conformal prediction method with the prediction set $\gC(x)$ and confidence level $1-\alpha$, $\alpha\in(0, 0.5)$, satisfies the following inequality:
    \begin{equation*} 
        H(Y|X)\leq h_b(\alpha) + \alpha\log(M) + \E([\log|\gC(x)|]^+),
    \end{equation*}
where $h_b(\cdot)$ is the binary entropy function, $[x]^+:=\max\{x,0\}$ and $H(Y|X)$ is computed using the true distribution $P_{XY}$. When the conformal prediction is merely based on the model output $\hat{Y}=f(X)$, the inequality can be modified to:
    \begin{equation*}
        H(Y|\hat{Y})\leq h_b(\alpha) + \alpha\log(M) + \E([\log|\gC(x)|]^+).
    \end{equation*}
\label{prop_app:fano_cp_hyx}
\end{proposition}
The proposition follows easily from Theorem \ref{thm:fano_list_decoding} by using the condition $\P(Y\in \gC(x))\geq 1-\alpha$. Note that in Theorem \ref{thm:fano_list_decoding}, we assume non-empty lists, whereas in Proposition~\ref{prop:fano_list_decoding} we allow empty prediction sets but apply the maximum operator $[x]^+:=\max\{x,0\}$. This is justified because the last term of Fano's inequality relates to the probability of correct assignments $Y \in \gC(X)$, which never happens for empty sets. See Proposition~\ref{prop:fano_maximal_set_size} for the proof, where the same result appears.

The bounds that we present in the main paper, model-based and simple Fano bounds, are actually derived through a slightly different root by leveraging the lower and upper bounds in the finite-sample guarantee of conformal prediction (Theorem~\ref{theo:main_cp}). We derive these other two bounds in Appendix~\ref{app:model_based_fano}.


\subsection{Related Work on Information Theoretic Inequalities}

The use of $f$-divergences goes back to works of Ali and Silvey, Csisz\'{a}r, and Morimoto in the 60s---see for instance \citep{ali_general_1966,csiszar_topological_1967,morimoto_markov_1963}. A key $f$-divergence inequality is the data processing inequality, which was used in information theory to establish various upper bounds on the achievability of coding schemes for different tasks---see \citep{sason_f_2016,sason_data-processing_2019} for an extensive survey. Moreover, the data processing inequality for $f$-divergences provides a unified way of obtaining many classical and new results, for example Fano's inequality \citep{fano_transmission_1949}. The tightness of data processing inequalities is discussed in terms of Bregman's divergence in \citep{collet_exact_2019, liese_divergences_2006} and in terms of $\chi^2$-divergence in \citep{sason_data-processing_2019}. 

When it comes to list decoding, there are a number of relevant inequalities in the literature. List decoding was introduced in the context of communication design by Elias \citep{elias_list_1957} and Wozencraft \citep{wozencraft_list_1958}. See also \citep{guruswami2004list} for a more recent overview of list decoding.
For fixed list size, the information theoretic bounds on list decoding were obtained in \citep{gallager_information_1986} using error exponent analysis. A generalization of Fano's inequality to list decoding was given in \citep{csiszar_information_2011} in the context of multi-user information theory, see also the general Fano inequality for list decoding presented in \citep{raginsky_concentration_2013}. For fixed list size, stronger inequalities, some based on Arimoto-R\'{e}nyi conditional entropy, were presented in \citep{sason_arimotorenyi_2017}. Variable size list decoding was discussed in \citep{sason_data-processing_2019} using the notion of  $E_\gamma$ resolvability first introduced in \citep{polyanskiy_channel_2010} related to the dependence testing bound. It was used again in the context of channel resolvability in \citep{liu_e__2017}, where some relevant inequalities have been obtained and discussed. A selection of the most relevant inequalities for list decoding can be found in \citep{sason_data-processing_2019}.

%% file: appendix/main_results.tex
\section{Proofs of Main Theoretical Results} \label{app:proofs}
In this section, we provide the proofs of our main results, namely the \emph{DPI bound} in Proposition~\ref{prop:dpi_bound}, the \emph{model-based Fano bound} in Proposition~\ref{prop:mbfano}, and the \emph{simple Fano bound} in Corollary~\ref{cor:simple_fano}. For notational convenience, we use the shorthand $\alpha_n = \alpha - \nicefrac{1}{n+1}$ in most of the steps of the derivations.
\subsection{DPI Bound} \label{app:dpi_proof}
We start with the DPI bound which we restate and proof below using the data processing inequalities discussed in Appendix~\ref{app:info_background}. Note that, when clear from the context, we remove explicit dependence on the calibration set $\gD_{cal}$ from the derivations. It is implicitly assumed that the probability of the event $Y\in \gC(x)$ is computed by marginalizing over $\gD_{cal}$.

\begin{repproposition}{prop:dpi_bound}
Consider any conformal prediction method with the prediction set $\gC(x)$ with the following finite sample guarantee:
\[
1-\alpha\leq \P(Y\in \gC(x))\leq 1-\alpha + \frac{1}{n+1}
\]
for any $\alpha\in(0, 0.5)$. For any arbitrary conditional distribution $Q_{Y|X}$, the true conditional distribution $P_{Y|X}$ and the input measure $P_X$, define the following two measures $Q:=P_XQ_{Y|X}$ and $P:=P_XP_{Y|X}$. We have for any $\alpha \in (0, 0.5)$
\begin{multline*}
    H(Y|X) \leq h_b(\alpha) + \left(1 -\alpha + \frac{1}{n + 1}\right)\log{Q}(Y \in \gC(x)) \\ + \alpha \log {Q}(Y \notin \gC(x)) - \E_{P_{XY}}\left[\log Q_{Y|X}\right].
\end{multline*}
\end{repproposition}
\begin{proof}
  Consider an arbitrary distribution $Q_{Y|X}$. Then we can use $P_{XY}$, and $P_X\times Q_{Y|X}$ in the data processing inequality for KL-divergence (Theorem~\ref{theo:dpi_cp}) to get:
\begin{equation} \label{eq:dpi_kl}
    D_{KL}(P_X P_{Y|X}||P_{X}Q_{Y|X}) \geq d_{KL}(P(Y \in \gC(x)) || Q(Y \in \gC(x)))
\end{equation}
Now note that we can decompose $D_{KL}(P_XP_{Y|X}||P_XQ_{Y|X})$ in terms of the conditional entropy $H(Y|X)$ and the cross-entropy $- \mathbb E_{P_{XY}}[\log Q_{Y|X}]:$
\begin{align*}
        D_{KL}(P_XP_{Y|X}||P_XQ_{Y|X}) &= \mathbb E_{P_{XY}}\left[\log \frac{P_XP_{Y|X}}{P_XQ_{Y|X}}\right] = \mathbb E_{P_{XY}}\left[\log \frac{P_{Y|X}}{Q_{Y|X}}\right] \\
        &= \mathbb E_{P_{XY}}[\log P_{Y|X}] - \mathbb E_{P_{XY}}[\log Q_{Y|X}] \\
        &= -H(Y|X) - \mathbb E_{P_{XY}}[\log Q_{Y|X}].
\end{align*}
With the decomposition above, we can rearrange the terms in (\ref{eq:dpi_kl}) to get the following upper bound on $H(Y|X)$
\begin{multline}
    -H(Y|X) - \E_{P_{XY}}\left[\log Q_{Y|X}\right] \geq d_{KL}(P(Y \in \gC(x)) || Q(Y \in \gC(x))) \\
    H(Y|X) \leq - d_{KL}(P(Y \in \gC(x)) || Q(Y \in \gC(x))) - \E_{P_{XY}}\left[\log Q_{Y|X}\right].
    \label{eq:entropy_kl_cross}
\end{multline}

We can then apply the upper and lower bounds from conformal prediction, i.e. $P(Y \in \gC(x)) \geq 1 - \alpha$ and $P(Y \notin \gC(x)) \geq \alpha_n$, to upper bound $d_{KL}(P(Y \in \gC(x)) || Q(Y \in \gC(x)))$ as follows.
\begin{multline*}
    - d_{KL}(P(Y \in \gC(x)) || Q(Y \in \gC(x))) = \\ h_b \left(P(Y\in \gC(x))\right) + 
    P(Y\in \gC(x)) \log Q(Y\in \gC(x)) + 
    P(Y\notin \gC(x)) \log Q(Y\notin \gC(x)) \\
    \leq h_b \left(\alpha \right) + \left(1 -\alpha \right)\log{Q}(Y \in \gC(x)) + \alpha_n \log {Q}(Y \notin \gC(x)),
\end{multline*}
where $h_b(\cdot)$ is the binary entropy function, that is, $h_b(\alpha)=-\alpha\log(\alpha) - (1-\alpha)\log(1-\alpha).$ The equality in the second line follows from the definition of the binary KL divergence, and we get the inequality simply by upper bounding each of the terms individually. In particular, note that $\log{Q}(Y \in \gC(x))$ and $\log {Q}(Y \notin \gC(x))$ are both negative, and $h_b(\cdot)$ is decreasing in $[0.5, 1.0]$ and symmetric about $0.5,$ such that for typical values of $\alpha < 0.5$
\begin{equation} \label{eq:bin_entropy}
    P(Y \in \gC(X)) \geq 1-\alpha \implies h_b \left(P(Y\in \gC(x))\right) \leq h_b(1-\alpha) = h_b(\alpha).
\end{equation}

Finally, we can replace the upper bound above into (\ref{eq:entropy_kl_cross}) to conclude the proof.
\begin{equation*}
    H(Y|X) \leq h_b(\alpha_n) + \left(1 -\alpha \right)\log{Q}(Y \in \gC(x)) + \alpha_n \log {Q}(Y \notin \gC(x)) - \E_{P_{XY}}\left[\log Q_{Y|X}\right].
\end{equation*}
  
\end{proof}

\begin{remark}
    The DPI bound always provides a tighter upper bound on the conditional entropy $H(Y|X)$ than the cross-entropy, which is easily verified in (\ref{eq:entropy_kl_cross}) since the KL term is always non-negative. This serves as an important motivation to optimize the DPI bound instead of the cross-entropy in conformal training.
\end{remark}

\begin{remark}
    The derivation of the DPI bound places no assumptions on the conditional distribution $Q_{Y|X}$. However, in practice, the underlying model $f$ often already provides such a distribution, and since it is typically trained to approximate $P_{Y|X}$ well, it makes sense to take $Q_{Y|X}$ as the distribution defined by the model itself. That is how we evaluate the DPI in all of our experiments. Finally, we can again use $H(Y|\hat{Y})$ instead of $H(Y|X)$ if the conformal method uses merely $\hat{Y}$. 
\end{remark}

\begin{remark}
    Typically, we have $\alpha \in (0.0, 0.5)$, and we use this fact in the proof to bound the binary entropy $h_b\left(P(Y\in \gC(x))\right).$ The same could have been done for $\alpha \in (0.5, 1.0)$, but since $1-\alpha$ lands in the increasing part of the binary entropy function between $0$ and $0.5$, we have to resort to the lower bound from conformal prediction to get
    \begin{equation*}
        P(Y \in \gC(X)) \leq 1-\alpha_n \implies h_b \left(P(Y\in \gC(x))\right) \leq h_b(1-\alpha_n) = h_b(\alpha_n).
    \end{equation*}
\end{remark}

\begin{remark}
    One of the appeals of the DPI bound is that the terms can be computed in a data-driven way using samples. While the cross-entropy can be estimated in an unbiased way with samples from the true distribution $P_{XY}$, we must be careful when estimating $Q(Y \in \gC(x))$. The main challenge is that $Q(Y \in \gC(x))$ appears inside a $\log$, and thus an empirical estimate $\hat{Q}(Y \in \gC(x))$ would yield a lower bound of the negative KL divergence and would be biased. We can get an upper confidence bound on this estimate via the empirical Bernstein inequality~\cite{maurer2009empirical}, which we restate below.

\begin{theorem}[Empirical Bernstein Inequality \cite{maurer2009empirical}]
Let $Z, Z_1, \dots, Z_n$ be i.i.d. random variables with values in [0, 1] and let $\delta > 0$. Then with probability at least $1 - \delta$ in the i.i.d. vector $\mathbf{Z} = (Z_1, \dots, Z_n)$ we have that
    \begin{equation*}
        \E[Z] - \frac{1}{n} \sum_{i=1}^n Z_i \leq \sqrt{\frac{2V_n(\mathbf{Z})\log(2/\delta)}{n}} + \frac{7\log(2/\delta)}{3 (n - 1)}, 
    \end{equation*}
where $V_n(\mathbf{Z})$ is the empirical variance over the $(Z_1, \dots, Z_n)$ samples.
\end{theorem}

By using this bound, we have with probability $1-\delta$:
    \begin{align*}
        Q(Y \in \gC(x)) & \leq \hat{Q}(Y \in \gC(x)) + \sqrt{\frac{2V_n(\mathbf{Z})\log(2/\delta)}{n}} + \frac{7\log(2/\delta)}{3 (n - 1)} := \tilde{Q}(Y \in \gC(x)),\\
        Q(Y \notin \gC(x)) &\leq \hat{Q}(Y \notin \gC(x)) + \sqrt{\frac{2V_n(\mathbf{Z})\log(2/\delta)}{n}} + \frac{7\log(2/\delta)}{3 (n - 1)} := \tilde{Q}(Y \notin \gC(x)),
    \end{align*}
where $Z_i = Q(y_i \in C(x_i))$ for $x_i$ sampled from $P_X$ and a prediction set $C(x_i)$ obtained with a calibration dataset sampled from $P_{\gD_{cal}}$. This yields the following inequality with probability $1-\delta$:
    \begin{equation*}
        H(Y|X) \leq h_b(\alpha) + \left(1 -\alpha \right)\log\tilde{Q}(Y \in \gC(x)) + \alpha_n \log \tilde{Q}(Y \notin \gC(x)) - {\E_{P_{XY}}\left[\log Q_{Y|X}\right]}.
    \end{equation*}
We can use this inequality to evaluate the bounds.
\end{remark}

\subsection{Model-Based Fano Bound} \label{app:model_based_fano}

In this section, prove the model-based Fano bound, which we restate in Proposition~\ref{prop:mbfano} below.

\begin{repproposition}{prop:mbfano}
Consider any conformal prediction method with the prediction set $\gC(x)$, and any distribution $Q$, with the following finite sample guarantee:
\[
1-\alpha\leq \P(Y\in \gC(x))\leq 1-\alpha + \frac{1}{n+1},
\]
for $\alpha\in(0, 0.5)$. Then, for the true distribution $P$, and for any probability distribution $Q$, we have
\begin{align*}
    H(Y|X) \leq  h_b(\alpha) &+ \alpha\E_{P_{Y, X, \gD_{cal}|Y \notin \gC(X)}}\left[- \log Q_{Y|X, \gC(x), Y \notin \gC(X)}\right] \\ &+ \left(1 - \alpha + \frac{1}{n+1}\right)\E_{P_{Y, X, \gD_{cal}|Y \in \gC(X)}}\left[- \log Q_{Y|X, \gC(x), Y \in \gC(X)}\right].
\end{align*}
\end{repproposition}
\begin{proof}
For notational convenience, define $E=\mathbf{1}(Y\in \gC(x))$, which by our assumption on the conformal prediction method means that $1- \alpha \leq \P(E = 1) \leq 1- \alpha_n$. The starting point is similar to the well-known proof of Fano's inequality:
\begin{align*}
    H(E,Y|X, \gD_{cal}) = H(Y|X, \gD_{cal}) + H(E|Y,X,\gD_{cal}) = H(Y|X, \gD_{cal}),
\end{align*}
where the last step follows because knowing $X,Y$ and $\gD_{cal}$, we know if $Y\in \gC(x)$, and therefore $H(E|Y,X, \gD_{cal})=0$. Furthermore, given the structure of the graphical model of the conformal prediction process (c.f. Figure~\ref{fig:graphical-model-cp}), we have that $H(Y|X) = H(Y|X, \gD_{cal})$, We now find an upper bound on $H(E,Y|X, \gD_{cal})$:
\begin{align}
    \nonumber H(E,Y|&X,\gD_{cal}) = H(E|X, \gD_{cal}) + H(Y|X,E, \gD_{cal}) \\
    \nonumber &= H(E|X, \gD_{cal}) +  P(E=0) H(Y|X,E=0,\gD_{cal})+  P(E=1) H(Y|X,E=1, \gD_{cal}) \\
    \nonumber &\leq H(E) +  P(E=0) H(Y|X,E=0,\gD_{cal})+  P(E=1) H(Y|X,E=1, \gD_{cal}) \\
    &\leq h_b(\alpha) +  P(E=0) H(Y|X,E=0,\gD_{cal})+  P(E=1) H(Y|X,E=1,\gD_{cal}),
\end{align}
where the first inequality follows from the fact that $H(E|X, \gD_{cal}) \leq H(E)$, and the last one comes from the same argument for $a < 0.5$ in (\ref{eq:bin_entropy}). We can continue as follows
\begin{align}
      \nonumber H(Y|X) & \leq h_b(\alpha) + P(E=0) H(Y|X,E=0,\gD_{cal})+  P(E=1) H(Y|X,E=1,\gD_{cal})\\
      \nonumber & = h_b(\alpha) + P(E=0)\E_{P_{Y, X, \gD_{cal}|E=0}}\left[- \log P_{Y|X, \gD_{cal}, E=0}\right]  \\
      \nonumber & \hskip0.095\textwidth + P(E=1)\E_{P_{Y, X, \gD_{cal}|E=1}}\left[- \log P_{Y|X, \gD_{cal}, E=1}\right]\\
      \nonumber & = h_b(\alpha) + P(E=0)\E_{P_{Y, X, \gD_{cal}|E=0}}\left[- \log Q_{Y|X, \gC(x), E=0}\right] \\
      \nonumber & \hskip0.095\textwidth - \E[D_{KL}(P_{Y|X, \gD_{cal}, E=0} || Q_{Y|X, \gC(x), E=0})] \\ 
      \nonumber & \hskip0.095\textwidth + P(E=1)\E_{P_{Y, X, \gD_{cal}|E=1}}\left[- \log Q_{Y|X, \gC(x), E=1}\right] \\
      \nonumber & \hskip0.095\textwidth - \E[D_{KL}(P_{Y|X, \gD_{cal}, E=1} || Q_{Y|X, \gC(x), E=1})]\\
      \nonumber &\leq h_b(\alpha) + P(E=0)\E_{P_{Y, X, \gD_{cal}|E=0}}\left[- \log Q_{Y|X, \gC(x), E=0}\right] \\ 
      & \hskip0.095\textwidth + P(E=1)\E_{P_{Y, X, \gD_{cal}|E=1}}\left[- \log Q_{Y|X, \gC(x), E=1}\right]
\end{align}
The first equality comes from the definition of the conditional entropy, whereas in the second equality we replace the true distribution $P_{Y|X}$ with an arbitrary conditional distribution $Q_{Y|X}$ plus the KL divergence between the two distributions. The last inequality then follows simply from the fact that the KL divergence is non-negative. Finally, we can leverage the finite-sample guarantees from conformal prediction, namely $P(E=0) \leq \alpha$ and $P(E=1) \leq 1 - \alpha_n$, to upper bound each term, which yields the proof. 

\begin{align} \label{eq:final_step_mb_fano}
    \nonumber H(Y|X) &\leq h_b(\alpha) + P(E=0)\E_{P_{Y, X, \gD_{cal}|E=0}}\left[- \log Q_{Y|X, \gC(x), E=0}\right] \\ 
    \nonumber & \hskip0.095\textwidth  + P(E=1)\E_{P_{Y, X, \gD_{cal}|E=1}}\left[- \log Q_{Y|X, \gC(x), E=1}\right] \\
    \nonumber &\leq h_b(\alpha) + \alpha\E_{P_{Y, X, \gD_{cal}|E=0}}\left[- \log Q_{Y|X, \gC(x), E=0}\right] \\
    & \hskip0.095\textwidth + \left(1 - \alpha_n \right)\E_{P_{Y, X, \gD_{cal}|E=1}}\left[- \log Q_{Y|X, \gC(x), E=1}\right]
\end{align}
\end{proof}

\begin{remark}
    When the conformal prediction is merely based on the model output $\hat{Y} = f(X)$, the model-based Fano bound can be modified to 
    \begin{align*}
        H(Y|\hat{Y}) \leq h_b(\alpha) &+ \alpha\E_{P_{Y, X, \gD_{cal}|Y \notin \gC(X)}}\left[- \log Q_{Y|X, \gC(x), Y \notin \gC(X)}\right] \\ &+ \left(1 - \alpha + \frac{1}{n+1}\right)\E_{P_{Y, X, \gD_{cal}|Y \in \gC(X)}}\left[- \log Q_{Y|X, \gC(x), Y \in \gC(X)}\right].
    \end{align*}
\end{remark}

\subsection{Simple Fano}
In the derivation of the model-based Fano bound above, we placed no assumptions on the distribution $Q_{Y|X}.$ One simple choice that we consider in this section is the uniform distribution $Q_{Y|X} = \nicefrac{1}{|\gY|}$, akin to the classical Fano's inequality \cite{fano_transmission_1949} and the list decoding result in Proposition~\ref{prop:fano_list_decoding}.
\begin{repcorollary}{cor:simple_fano}
Consider any conformal prediction method with the prediction set $\gC(x)$, and any distribution $Q$, with the following finite sample guarantee:
\[
1-\alpha\leq \P(Y\in \gC(x))\leq 1-\alpha + \frac{1}{n+1}.
\]
For $\alpha\in(0, 0.5)$ we have the following inequality:
\begin{multline*}
    H(Y|X) \leq h_b(\alpha) + \alpha\E_{P_{Y, X, \gD_{cal}|Y \notin \gC(X)}}\left[\log (|\gY| - |C(X)|)\right] \\ + \left(1 - \alpha_n\right)\E_{P_{Y, X, \gD_{cal}|Y \in \gC(X)}}\left[ \log |C(X)|\right].
\end{multline*}
\end{repcorollary}

\begin{proof}
    Note that if we make an error ($E=0$) the correct class will not be inside $\gC(X)$ and since $Q_{Y|X}$ is uniform the probability will be spread equally among the remaining $|\gY| - |\gC(X)|$ labels, and we have $Q_{Y|X, \gC(x), E=0} = \frac{1}{|\gY| - |\gC(X)|}$. Through the same logic, we get that $Q_{Y|X, \gC(x), E=1} = \frac{1}{|\gC(X)|},$ and plugging both into (\ref{eq:final_step_mb_fano}) we get the simple Fano bound
\end{proof}

\begin{remark}
Once again, when the conformal prediction prediction is merely based on the model output $\hat{Y} = f(X)$, the inequality can be modified to
    \begin{multline*}
    H(Y|\hat Y) \leq h_b(\alpha) + \alpha\E_{P_{Y, X, \gD_{cal}|Y \notin \gC(X)}}\left[\log (|\gY| - |C(X)|)\right] \\ + \left(1 - \alpha_n\right)\E_{P_{Y, X, \gD_{cal}|Y \in \gC(X)}}\left[ \log |C(X)|\right].
\end{multline*}
\end{remark}

%% file: appendix/list_decoding.tex
\section{Further Theoretical Results and Proofs on the Prediction Set Size}
\label{app:set_size}

\subsection{Fano's inequality for maximal prediction set size}
If we leverage the upper bound on the prediction set size, we can find a lower bound on the maximum coverage set size. 
\begin{proposition} \label{prop:fano_maximal_set_size}
Suppose that $|\gY|=M$. Consider any conformal prediction method that constructs the prediction set $\gC(X)$ with the following finite-sample guarantee:
\[
1 - \alpha \leq \P(Y\in \gC(X))\leq 1-\alpha + \frac{1}{n+1}
\]
for $\alpha\in(0, 0.5)$. Then, we have the following inequality:
\begin{align*}
    H(Y|X)&\leq h_b\left(\alpha\right) + \alpha\log M + (1-\alpha + \frac{1}{n+1})\sup_{\gD_{cal}}\sup_{x\in\supp(P_X)}\log|\gC(x)|.
\end{align*}
We can similarly replace $H(Y|X)$ with $H(Y|\hat{Y})$.
\end{proposition}
\begin{proof}
    We use the conditional data processing inequality with $f(x)=x\log x$, $\mathrm{P}=P_{\gD_{cal}XY}$, and $\mathrm{Q}=P_{\gD_{cal}X}\times U_M$ where $U_M$ is the uniform distribution over $\gY$. We fix the input to $X=x$, and the calibration set $\gD_{cal}$. The conditional $f$-divergence, conditioned on $\gD_{cal}$ and $X$, is given by:
\begin{align*}
   D_f(\mathrm{Q}_{Y|X=x,\gD_{cal}}||\mathrm{P}_{Y|X=x,\gD_{cal}}) &= D_{KL}(P_{Y|\gD_{cal}X=x}||U_M) \\ &= \log M -H(Y|X=x),
\end{align*}
where the last step follows from the independence of the calibration set $\gD_{cal}$ and $(X,Y)$.
This means that
\begin{equation} \label{eq:exp_df}
    \E_{\gD_{cal},X} D_f(P_{Y|X=x,\gD_{cal}}||Q_{Y|X=x,\gD_{cal}}) = \log M - H(Y|X).
\end{equation}
On the other hand,  we have
\begin{align} \label{eq:df}
    \nonumber d_f&\left( Q(Y\in \gC(x)|\gD_{cal},X=x) || P(Y\in \gC(x)|\gD_{cal},X=x)\right) \\
    \nonumber & = d_{KL}\left( P(Y\in \gC(x)|\gD_{cal},X=x) || Q(Y\in \gC(x)|\gD_{cal},X=x) \right) \\
    \nonumber & = - h_b \left(P(Y\in \gC(x)|\gD_{cal},X=x)\right) - P(Y\in \gC(x)|\gD_{cal},X=x) \log \frac{|\gC(x)|}{M} \\
    & \quad -
    P(Y\notin \gC(x)|\gD_{cal},X=x) \log \frac{M-|\gC(x)|}{M}.
\end{align}
Now, we can plug both (\ref{eq:exp_df}) and (\ref{eq:df}) into the data processing inequality of Theorem~\ref{theo:cond_dpi_cp}. Rearranging the terms and getting the expectation from both sides w.r.t. $\gD_{cal}$ and $X$ would yield:
\begin{align*}
    H(Y|X)&\leq \E [h_b\left(P(Y\in \gC(x)|\gD_{cal},X)\right)] \\ 
    & \quad + \E [P(Y\notin \gC(x)|\gD_{cal},X)\log({M-|\gC(x)|})] + \E[P(Y\in \gC(x)|\gD_{cal},X) \log {|\gC(x)|} ]\\
    & \quad - \E [P(Y\notin \gC(x)|\gD_{cal},X) \log M] - \E[P(Y\in \gC(x)|\gD_{cal},X)\log M] + \log M  \\
    &\leq h_b\left(P(Y\in \gC(x))\right) + P(Y\notin \gC(x))\log M + \E[P(Y\in \gC(x)|\gD_{cal},X) \log {|\gC(x)|} ]\\
    &\leq h_b\left(\alpha\right) + \alpha\log M + \E[P(Y\in \gC(x)|\gD_{cal},X) \log {|\gC(x)|} ].
\end{align*}
Note that the $\log M$ terms in the third line cancel each other out, and that the second inequality comes from resolving the expectations and $\log(M - |\gC(X)|) \leq \log M.$ Further, in the last inequality, we used the concavity of binary entropy as in (\ref{eq:bin_entropy}).
Finally, using respectively the inequalities $P(Y\in \gC(x)|\gD_{cal},X) \leq 1$ and $\log|\gC(x)|\leq \sup_{\gD_{cal}}\sup_{x\in\gX}\log|\gC(x)|$, we get two inequalities:
\begin{align*}
    H(Y|X)&\leq h_b\left(\alpha\right) + \alpha\log M + \E\left([\log {|\gC(x)|}]^+\right)\\
    H(Y|X)&\leq h_b\left(\alpha\right) + \alpha\log M + (1-\alpha + \frac{1}{n+1})\sup_{\gD_{cal}}\sup_{x\in\supp(P_X)}\log|\gC(x)|.
\end{align*}
The first one is the Fano's inequality for list decoding from Proposition~\ref{prop:fano_list_decoding} while the second one yields the theorem. Note that if $\gC(x)$ is an empty set, the probability $P(Y\in \gC(x)|\gD_{cal},X)$ is zero, and the term $P(Y\in \gC(x)|\gD_{cal},X) \log {|\gC(x)|}$ disappears. Therefore, if we use the inequality $P(Y\in \gC(x)|\sup_{\gD_{cal}},X) \leq 1$, we need to introduce the term $[\log {|\gC(x)|}]^+$  to keep the expectation well defined.
\end{proof}

\subsection{Fano's inequality for lower bound on prediction set size}
In a similar manner, we can also obtain lower bounds for the set size. More specifically, we have that 
\begin{align}
    Q_{Y|X, \gC(x), E=1} & = \frac{q(y|x)\mathbb{I}[y \in \gC(x)]}{\sum_{y \in \gC(x)} q(y|x)} = \frac{1}{|\gC(x)|\E_{u(y_{\gC(x)})}[q(y|x)]}q(y|x)\mathbb{I}[y \in \gC(x)] 
    \nonumber \\ & := \frac{1}{|\gC(x)|\E_{u(y_{\gC(x)})}[q(y|x)]}\hat{Q}^1_{Y|X} \label{def:hat_q_1} \\
    Q_{Y|X, \gC(x), E=0} & = \frac{q(y|x)\mathbb{I}[y \notin \gC(x)]}{\sum_{y \notin \gC(x)} q(y|x)} = \frac{1}{(M- |\gC(x)|)\E_{u(y_{\bar{\gC(x)}})}[q(y|x)]}q(y|x)\mathbb{I}[y \notin \gC(x)] 
    \nonumber \\
    & := \frac{1}{(M - |\gC(x)|)\E_{u(y_{\bar{\gC(x)}})}[q(y|x)]}\hat{Q}^0_{Y|X} \label{def:hat_q_0}
\end{align}
where $u(y_{\gC(x)})$ and $u(y_{\bar{\gC(x)}})$ denote uniform distributions over the labels in the confidence set. By considering the standard conformal prediction bounds on the error probabilities, we  have that 
\begin{align*}
    H(Y|X) & \leq h_b(\alpha) + \alpha\E_{P_{Y, X, \gD_{cal}|E=0}}\left[- \log \hat{Q}^0_{Y|X} + \log(M- \gC(x)) + \log\E_{u(y_{\bar{\gC(x)}})}[q(y|x)] \right] \nonumber \\ 
    \nonumber & + \paran{1 - \alpha + \frac{1}{n + 1}}\E_{P_{Y, X, \gD_{cal}|E=1}}\left[- \log \hat{Q}^1_{Y|X} + \log |\gC(x)| + \log\E_{u(y_{\gC(x)})}[q(y|x)] \right] \\
    & \leq h_b(\alpha) + \alpha\log M + \alpha\E_{P_{Y, X, \gD_{cal}|E=0}}\left[- \log \hat{Q}^0_{Y|X} + \log\E_{u(y_{\bar{\gC(x)}})}[q(y|x)]\right] \nonumber \\ 
    & + \left(1 - \alpha + \frac{1}{n+1}\right)\E_{P_{Y, X, \gD_{cal}|E=1}}\left[- \log \hat{Q}^1_{Y|X} + \log |\gC(x)| + \log\E_{u(y_{\gC(x)})}[q(y|x)] \right]
\end{align*}
which leads to
\begin{align*}
    \nonumber \E_{E=1}&[\log |\gC(x)|] \geq \\ &\frac{H(Y|X) - h_b(\alpha) - \alpha\log M - \alpha\E_{P_{Y, X, \gD_{cal}|E=0}}\left[- \log \hat{Q}^0_{Y|X} + \log\E_{u(y_{\bar{\gC(x)}})}[q(y|x)]\right]}{1 - \alpha + \frac{1}{n + 1}} \nonumber \\ & \quad - \E_{P_{Y, X, \gD_{cal}|E=1}}\left[- \log \hat{Q}^1_{Y|X} + \log\E_{u(y_{\gC(x)})}[q(y|x)] \right].
\end{align*}

Note that when $E=1$, we know that $Y\in \gC(x)$, and therefore $|\gC(x)|>0$ and $[\log|\gC(x)|]^+=\log|\gC(x)|$. Using this, we can find an upper bound on $\E_{E=1}(\log (|\gC(x)|))$ as follows:
\begin{align*}
    \nonumber \E_{E=1}(\log (|\gC(x)|)) & = E_{E=1}([\log (|\gC(x)|)]^+)\\
    \nonumber &= \left(\frac{\E([\log |\gC(x)|]^+)}{P(E=1)} - \frac{P(E=0)}{P(E=1)}\E_{E=0}([\log |\gC(x)|]^+)\right) \\
    & \leq \left(\frac{\E([\log |\gC(x)|]^+)}{P(E=1)}\right) 
    \leq  \left(\frac{\E([\log |\gC(x)|]^+)}{1 - \alpha}\right).
\end{align*}
This leads to:
\begin{align*}
    \nonumber \E&([\log|\gC(x)|]^+) \geq \\ &(1-\alpha)\frac{H(Y|X) - h_b(\alpha) - \alpha\log M - \alpha\E_{P_{Y, X, \gD_{cal}|E=0}}\left[- \log \hat{Q}^0_{Y|X} + \log\E_{u(y_{\bar{\gC(x)}})}[q(y|x)]\right]}{1 - \alpha + \frac{1}{n+1}} \nonumber \\ & - (1-\alpha)\E_{P_{Y, X, \gD_{cal}|E=1}}\left[- \log \hat{Q}^1_{Y|X} + \log\E_{u(y_{\gC(x)})}[q(y|x)] \right] 
\end{align*}
All the above terms can be approximated from samples. We summarize this in the following proposition.
\begin{proposition}
    For any conformal prediction scheme with the coverage guarantee of $1-\alpha$, and any distribution $q(\cdot)$, we have:
    \begin{align}
    \nonumber &\E([\log|\gC(x)|]^+) \geq \\
    \nonumber & (1-\alpha)\frac{H(Y|X) - h_b(a) - a\log M - \alpha\E_{P_{Y, X, \gD_{cal}|E=0}}\left[- \log \hat{Q}^0_{Y|X} + \log\E_{u(y_{\bar{\gC(x)}})}[q(y|x)]\right]}{1 - \alpha + \frac{1}{n+1}} \nonumber \\ 
    & - (1-\alpha)\E_{P_{Y, X, \gD_{cal}|E=1}}\left[- \log \hat{Q}^1_{Y|X} + \log\E_{u(y_{\gC(x)})}[q(y|x)] \right] 
\end{align}
where $\hat{Q}^0_{Y|X} = q(y|x)\mathbb{I}[y \notin \gC(x)]$ and $\hat{Q}^1_{Y|X} = q(y|x)\mathbb{I}[y \in \gC(x)]$. When the conformal prediction is merely based on the model output $\hat{Y} = f(X)$, the inequality can be modified to 
\begin{align}
    \nonumber &\E([\log|\gC(x)|]^+) \geq \\
    \nonumber & (1-\alpha)\frac{H(Y|\hat{Y}) - h_b(a) - a\log M - \alpha\E_{P_{Y, X, \gD_{cal}|E=0}}\left[- \log \hat{Q}^0_{Y|X} + \log\E_{u(y_{\bar{\gC(x)}})}[q(y|x)]\right]}{1 - \alpha + \frac{1}{n+1}} \nonumber \\ & \quad - (1-\alpha)\E_{P_{Y, X, \gD_{cal}|E=1}}\left[- \log \hat{Q}^1_{Y|X} + \log\E_{u(y_{\gC(x)})}[q(y|x)] \right] 
\end{align}
\label{prop:coverage_set_size_model_based_fano}
\end{proposition}

\begin{remark}
    If we use the uniform distribution in the above bound, we get a bound similar to what is obtained from Fano's inequality given in Proposition \ref{prop_app:fano_cp_hyx}, but with an additional factor of $\frac{1-\alpha}{1-\alpha+\frac{1}{n+1}}$. Since the factor is smaller than one, the current bound with the choice of uniform distribution is looser than Fano's bound, although the gap vanishes for large $n$. 
\end{remark}

%% file: appendix/conftr.tex
\clearpage
\section{Conformal Training} \label{app:conftr}
Split conformal prediction (SCP) \cite{papadopoulos2002inductive} has quickly become a popular framework for uncertainty quantification, largely thanks to its computational efficiency. One only needs access to a separate calibration data set to derive prediction sets with valid marginal coverage from any pretrained model. Given that training new machine learning models is becoming ever more time-consuming and expensive with new, larger architectures, this ability to apply conformal prediction to existing models is invaluable in a number of applications. Yet, it is reasonable to expect that the performance of the final set predictor could be improved if the conformal prediction process were to be accounted for during training of the model as well, steering the model towards better predictive efficiency, as it were. That is the motivation behind the line of work that we broadly refer to as \emph{conformal training} \cite{colombo2020training,bellotti2021optimized,stutz2022learning}. In a nutshell, conformal training introduces a differentiable, and hence approximate, conformal prediction step during training so that one can directly optimize for desired properties of the set predictor, most notably its predictive efficiency. In what follows we give an overview of conformal training, focusing on classification tasks.

The idea of conformal training has been proposed concomitantly in \cite{bellotti2021optimized,colombo2020training}. Here we follow the approach in \citep{bellotti2021optimized}, where the key idea is to relax the prediction set defined by the model $f$ and define ``soft'' prediction sets $\hat \gC_f(x)$, which contain each of the labels $y \in \gY$ with a certain probability. That is, if $\gC_f(x,y) \in \{0, 1\}$ is the hard assignment of label $y$ to $\gC(x)$, a corresponding soft version of this assignment can be defined as 
\begin{equation} \label{eq:soft_c}
    \hat \gC_f(x, y) := \sigma \left(\frac{\hat q - s_f(x, y)}{T}\right)
\end{equation}
where $s_f(x, y)$ is the non-conformity score function defined by model $f$ evaluated at $(x, y)$, $\hat q$ is a thresholding value, $\sigma$ is the logistic sigmoid function, and $T$ is a temperature parameter controlling the smoothness of the soft assignment. Then we can define $\hat \gC_f(x)$ as the vector collection of all the soft assignments $\hat \gC_f(x, y)$ for all labels $y \in \gY$. Similarly the size of $\hat \gC_f(x)$ can be naturally defined as
\begin{equation*} 
    |\hat \gC_f(x)| := \sum_{y \in \gY} \hat \gC_f(x, y).
\end{equation*}
\citet{bellotti2021optimized} then proposes the following loss functions which are computed for each training batch $\gB$
\begin{equation*}
    \gL_{size}(f)  = \frac{1}{|\gB|} \sum_{x \in \gB} g\left(|\hat \gC_f(x))|\right) \,\,\,
    \gL_{coverage}(f) = \left( \left[\frac{1}{|\gB|}\sum_{(x, y) \in \gB} \hat \gC_f(x, y)\right] - (1-\alpha)\right)^2,
\end{equation*}
where $\alpha$ is the desired coverage rate and $g$ is a user-defined function of the prediction set size, e.g. the log function. Intuitively, $\gL_{size}$ encourages small (efficient) prediction sets, whereas $\gL_{coverage}$ penalizes deviations from the target coverage of $1-\alpha$. Naturally, there is a trade-off between these two objectives, inefficiency and coverage, so both loss terms are optimized together with a hyperparameter $\lambda$ governing the influence of each term:
\begin{align} \label{eq:bel_loss}
    \gL(f) = \gL_{size}(f) &+ \lambda \gL_{coverage}(f).
\end{align}
Importantly, \citet{bellotti2021optimized} argues the choice of the threshold $\hat q$ in (\ref{eq:soft_c}) is immaterial since the model can learn to shift its outputs (in logit space) accordingly to match the constraints in $\gL_{coverage}(f)$. We can then directly optimize (\ref{eq:bel_loss}) via stochastic gradient descent methods during training since it is fully differentiable with respect to the model parameters.

\citet{stutz2022learning} build on the work of \citet{bellotti2021optimized} by noticing that the calibration step is an important component in conformal prediction that should also be accounted for during training. To that end, they propose to split each training batch $\gB$ in two: the $\gB_{cal}$ half used for calibration, and the $\gB_{test}$ used for testing. Now, instead of using an arbitrary threshold $\hat q$, we compute it using the quantile of $\gB_{cal}$, or concretely
\begin{equation*} 
    \hat q = \quant(1-\alpha; \{s_f(x, y) : (x, y) \in \gB_{cal}\})
\end{equation*}
With this modification, we no longer need to enforce valid coverage via $\gL_{coverage}(f)$ and can optimize for low inefficiency directly by minimizing $\gL_{size}(f)$ on $\gB_{test}$. In that case, however, we only get a learning signal from $\gB_{test}$, since the quantile operation applied to $\gB_{cal}$ is non-differentiable. \citet{stutz2022learning} bypass that limitation via differentiable sorting operators \cite{cuturi2019differentiable,blondel2020fast,petersen2022monotonic}, in particular via a version of differentiable sorting networks. In our experiments, we considered both fast sort \cite{blondel2020fast} and the monotonic differentiable sorting networks of \cite{petersen2022monotonic} but finally chose the latter since they proved more stable and provided richer gradient signals.

This version of their approach, which we refer to as ConfTr, only optimizes the size loss, but \citet{stutz2022learning} also proposed another variant which includes a classification loss term as follows
\begin{equation} \label{eq:cftr_class}
    \gL_{class}(f) = \frac{1}{|\gB|} \sum_{(x,y) \in \gB} \sum_{\hat y \in \gY} L_{y, \hat y} \left[ \left(1-\hat\gC_f(x, y) \right)\delta[\hat y = y] + \hat\gC_f(x, y)\delta[\hat y \neq y] \right],
\end{equation}
where $\delta$ is the indicator function, and $L$ is a user-defined  square matrix of size $|\gY|^2$ with $L_{y, \hat y}$ capturing some similarity notion between $y$ and $\hat y$. In our experiments, as well as most experiments in the original paper \cite{stutz2022learning}, no prior information about the classification problem is assumed, in which case $L$ is taken to be the identity matrix of size $|\gY|$. Therefore, we have two variants of conformal training as proposed in \cite{stutz2022learning}: \textbf{ConfTr} that optimizes only $\gL_{size}$, and \textbf{ConfTr$_{\text{class}}$} that jointly optimizes $\gL_{size}$ and $\gL_{class}$, both of which are included in our experiments.

Our own approach to conformal training follows the same recipe from \cite{stutz2022learning}, i.e., we also simulate a split conformal prediction step during training by splitting each training batch into two and using differentiable sorting operators (see Algorithm~\ref{alg:conftr}). The key difference is in how we define the training objectives, which we derive from first principles and standard information theory inequalities. Not only do our upper bounds, DPI (\ref{eq:dpi_bound}), MB Fano (\ref{eq:mb_fano_bound}) and Fano (\ref{prop:simple_fano}), outperform the ConfTr objectives in many cases, but they also do away with a few hyperparameters. Namely, the function $g$ in $\gL_{size}$, and hyperparameters controlling the relative importance of $\gL_{size}$ and $\gL_{class}$.

\subsection{Deriving Conformal Training from Fano's bound} \label{app:conftr_fano}
Through Proposition~\ref{prop:mbfano},
we can connect Fano's bound for list decoding to the size loss from \cite{stutz2022learning}, proposed for conformal training. Assuming a uniform distribution for $Q$ we can show that
\begin{align*}
    H(Y|X) & \leq h_b(\alpha) + \alpha\E_{E=0}\left[\log (|Y| - |\gC(x)|)\right] 
+ \left(1 - \alpha_n \right)\E_{E=1}\left[ \log |\gC(x)|\right]\\
    & \leq h_b(\alpha) + \alpha \log|Y| + \left(1 - \alpha_n \right) \E_{E=1}\left[\log |\gC(x)|\right]\\
    & \leq h_b(\alpha) + \alpha \log|Y| + \left(1 - \alpha_n \right) \log \E_{E=1}\left[|\gC(x)|\right]\\
    & \leq h_b(\alpha) + \alpha \log|Y| - \left(1 - \alpha_n \right) \log(1 - \alpha) + \left(1 - \alpha_n \right) \log \E\left[|\gC(x)|\right]. 
\end{align*}
Note that in the first line we have the simple Fano bound, whereas in the last one we have the ConfTr objective, namely $\log \E\left[|\gC(x)|\right]$, multiplied by $1-\alpha_n$ plus a constant that depends only on $\alpha.$ Therefore, we ground ConfTr as minimizing a looser upper bound to the true conditional entropy of the data than the simple Fano bound we provide in Corollary~\ref{cor:simple_fano}. Moreover, the simple Fano bound can be further improved with an appropriate choice of $Q$, for instance as given by the model $f$, in the model-based Fano bound of Proposition~\ref{prop:mbfano}.

%% file: appendix/experiments.tex
\clearpage
\section{Experiments} \label{app:experiments}
In this section, we present further experimental results for conformal training in the centralized and federated setting. We start by defining the splits and architectures used for each data set, which are listed in Table~\ref{tab:exp_settings}. In most aspects, we follow the experimental design of \citep{stutz2022learning}. All experiments were conducted on commercially available NVIDIA GPUs using our own implementation in Python 3 and Pytorch \citep{paszke2017automatic}, which can be found at \href{https://github.com/Qualcomm-AI-research/info_cp}{\nolinkurl{github.com/Qualcomm-AI-research/info_cp}}. All data sets were retrieved directly from \texttt{torchvision} \citep{torchvision2016}.

\begin{table}[!h]
    \centering
    \caption{Experimental settings for each data set, with $|\gD_{train}|$, $|\gD_{cal}|$ and $|\gD_{test}|$ the sizes of train, calibration and test splits, respectively.}
    \label{tab:exp_settings}
    \begin{tabular}{c|c c c c c}
        \toprule
        Data set & $|\gD_{train}|$ & $|\gD_{cal}|$ & $|\gD_{test}|$ & Epochs & Architecture \\
        \midrule
        MNIST \cite{lecun1998gradient} & 55K & 5K & 10K & 50 & 1-layer MLP \\
        Fashion-MNIST \cite{xiao2017fashion} & 55K & 5K & 10K & 150 & 2-layer MLP \\
        EMNIST \cite{cohen2017emnist} & 628K & 70K & 116K & 75 & 2-layer MLP \\
        CIFAR10 \cite{krizhevsky2009learning} & 45K & 5K & 10K & 150 & ResNet-34 \\
        CIFAR100 \cite{krizhevsky2009learning} & 45K & 5K & 10K & 150 & ResNet-50 \\
        \bottomrule
    \end{tabular}
\end{table}

Regarding the architectures, we also closely follow the experimental setup in \cite{stutz2022learning}. For MNIST we have a simple linear model, whereas for Fashion-MNIST and EMNIST we use 2-layer MLPs with 64 and 128 hidden units for first and second layers, respectively. For the CIFAR data sets, we use the default ResNet implementations from \emph{torchvision} \cite{torchvision2016}, but changing the first convolution to have a kernel size of 3 and unitary stride and padding. We use Pytorch's default weight initialization strategy for all architectures. For all datasets, we use a regular SGD optimizer with momentum $0.9$ and Nesterov gradients, accompanied by a step scheduler multiplying the initial learning rate by $0.1$ after $2/5$, $3/5$ and $4/5$ of the total number of epochs. We only use data augmentations on the CIFAR datasets, and differently from \cite{stutz2022learning}, we only apply random flipping and cropping for both CIFAR10 and CIFAR100.

\subsection{Centralized Setting}
We followed the experimental procedure of \cite{stutz2022learning}, and for each dataset and each method, we ran a grid search over the following hyperparameters using \emph{ray tune} \cite{moritz2018ray}: 
\begin{itemize}
    \item \textbf{Batch size} with possible values in $\{100, 500, 1000\}$.
    \item \textbf{Learning rate} with possible values in $\{0.05, 0.01, 0.005\}$.
    \item \textbf{Temperature} used in relaxing the construction of prediction sets at training time. We considered temperature values in $\{0.01, 0.1, 0.5, 1.0\}$.
    \item \textbf{Steepness} of the differentiable sorting algorithm (monotonic sorting networks with Cauchy distribution \cite{petersen2022monotonic}), which regulates the smoothness of the sorting operator; the higher the steepness value, the closer the differentiable sorting operator is to standard sorting. We considered steepness values in $\{1, 10, 100\}$.
\end{itemize}

In Tables~\ref{tab:hyper_mnist}, \ref{tab:hyper_fashion}, \ref{tab:hyper_emnist}, \ref{tab:hyper_cifar10} and \ref{tab:hyper_cifar100}, we report the best hyperparameters found for each dataset and method as well as the average prediction set size for threshold CP \cite{sadinle2019least} computed in the probability domain (THR) and APS \cite{romano2020classification}, as well as the test accuracy. 
Importantly, similarly to \cite{stutz2022learning}, in all cases we only train the models to optimize threshold CP with log-probabilities. We confirm the observation in \cite{stutz2022learning} that other methods, and notably APS, are unstable during training, probably because it forces us to operate in the probability domain, as opposed to the more optimization-friendly logits or log-probabilities. Nevertheless, we still select hyperparameters according to the best performance with respect to each CP method, and that is why we have different optimal hyperparameters for THR and APS for each data set and each conformal training objective. We note ConfTr and ConfTr$_{\text{class}}$ require extra hyperparameters like the target size and weights attributed to each loss term (see Appendix~\ref{app:conftr}). For those hyperparameters, we use the best values for each data set as reported in \cite{stutz2022learning}.

As described in the main paper, we use the default train and test splits of each data set but transfer 10\% of the training data to the test data set. We train the classifiers only on the remaining 90\% of the training data and, at test time, run SCP with 10 different calibration/test splits by randomly splitting the enlarged test data set. All results reported in the paper are given by the average ($\pm$ one standard deviation) computed across these 10 random splits. Crucially, to avoid overfitting to the test data, the grid search was done solely on the 90\% of the training data not used for testing.

\begin{table*}[!h]
    \caption{Hyperparameter Search for MNIST.}
    \centering
    \label{tab:hyper_mnist}
    \resizebox{\textwidth}{!}{
    \begin{tabular}{c c | c c c c | c c | c}
        Bound & Optimized for & batch size & lr & temperature & steepness & THR & APS & Test Acc.\\
        \midrule
        \multirow{ 2}{*}{CE} & THR & 100 & 0.01 & - & - & $2.29_{\pm 0.18}$ & $2.50_{\pm 0.08}$ & 0.93 \\
        & APS & 500 & 0.05 & - & - & $2.28_{\pm 0.19}$ & $2.50_{\pm 0.08}$ & 0.93 \\
        \cmidrule[0.5pt](l){1-9}
        \multirow{ 2}{*}{ConfTr} & THR & 500 & 0.05 & 0.1 & 100 & $6.28_{\pm 0.71}$ & $9.81_{\pm 0.06}$ & 0.90 \\
        & APS & 500 & 0.01 & 1.0 & 100 & $2.08_{\pm 0.10}$ & $2.10_{\pm 0.07}$ & 0.90 \\
        \cmidrule[0.5pt](l){1-9}
        \multirow{ 2}{*}{ConfTr-class} & THR & 500 & 0.005 & 1.0 & 10 & $2.09_{\pm 0.11}$ & $2.15_{\pm 0.12}$ & 0.91 \\
        & APS & 500 & 0.01 & 0.1 & 100 & $2.12_{\pm 0.11}$ & $2.13_{\pm 0.13}$ & 0.90 \\
        \cmidrule[0.5pt](l){1-9}
        \multirow{ 2}{*}{Fano} & THR & 100 & 0.005 & 0.1 & 100 & $2.09_{\pm 0.12}$ & $2.12_{\pm 0.09}$ & 0.91 \\
        & APS & 100 & 0.01 & 0.5 & 100 & $2.11_{\pm 0.08}$ & $2.12_{\pm 0.08}$ & 0.91 \\
        \cmidrule[0.5pt](l){1-9}
        \multirow{ 2}{*}{MB Fano} & THR & 1000 & 0.005 & 0.1 & 100 & $2.24_{\pm 0.12}$ & $2.96_{\pm  0.11}$ & 0.91 \\
        & APS & 100 & 0.005 & 0.1 & 10 & $2.43_{\pm 0.22}$ & $2.49_{\pm 0.19}$ & 0.92 \\
        \cmidrule[0.5pt](l){1-9}
        \multirow{ 2}{*}{DPI } & THR & 1000 & 0.005 & 0.01 & 100 & $2.24_{\pm 0.17}$ & $2.87_{\pm 0.12}$ & 0.92 \\
        & APS & 500 & 0.05 & 0.1 & 100 & $2.29_{\pm 0.17}$& $2.64_{\pm 0.07}$ & 0.92 \\
        \bottomrule
    \end{tabular}
    }
\end{table*}
\begin{table*}[!h]
    \caption{Hyperparameter Search for Fashion-MNIST.}
    \centering
    \label{tab:hyper_fashion}
    \resizebox{\textwidth}{!}{  
    \begin{tabular}{c c | c c c c | c c | c}
    \toprule
        Bound & Optimized for & batch size & lr & temperature & steepness & THR & APS & Test Acc. \\
        \midrule
        \multirow{ 2}{*}{CE} & THR & 1000 & 0.005 & - & - & $2.39_{\pm 0.13}$ & $2.74_{\pm 0.20}$ & 0.87 \\
        & APS & 100 & 0.05 & - & - & $2.18_{\pm 0.17}$ & $2.41_{\pm 0.17}$ & 0.89 \\
        \cmidrule[0.5pt](l){1-9}
        \multirow{ 2}{*}{ConfTr} & THR & 100 & 0.01 & 1.0 & 1.0 & $1.73_{\pm 0.06}$ & $1.89_{\pm 0.09}$ & 0.89 \\
        & APS & 100 & 0.01 & 1.0 & 1.0 & $1.73_{\pm 0.06}$ & $1.89_{\pm 0.09}$ & 0.89 \\
        \cmidrule[0.5pt](l){1-9}
        \multirow{ 2}{*}{ConfTr-class} & THR & 100 & 0.01 & 1.0 & 10.0 & $5.11_{\pm 0.49}$ & $8.28_{\pm 3.17}$ & 0.88 \\
        & APS & 500 & 0.005 & 1.0 & 100.0 & $1.79_{\pm 0.06}$ & $1.79_{\pm 0.07}$ & 0.88 \\
        \cmidrule[0.5pt](l){1-9}
        \multirow{ 2}{*}{Fano} & THR & 100 & 0.05 & 0.5 & 100.0 & $1.70_{\pm 0.05}$ & $2.06_{\pm 0.08}$ & 0.88 \\
        & APS & 100 & 0.01 & 1.0 & 100.0 & $1.76_{\pm 0.04}$ & $1.87_{\pm 0.05}$ & 0.87 \\
        \cmidrule[0.5pt](l){1-9}
        \multirow{ 2}{*}{MB Fano} & THR & 100 & 0.05 & 0.5 & 10.0 & $1.80_{\pm 0.08}$ & $2.71_{\pm 0.14}$ & 0.89 \\
        & APS & 100 & 0.05 & 1.0 & 100.0 & $1.84_{\pm 0.10}$ & $2.25_{\pm 0.14}$ & 0.89\\
        \cmidrule[0.5pt](l){1-9}
        \multirow{ 2}{*}{DPI } & THR & 100 & 0.05 & 0.01 & 10.0 & $1.73_{\pm 0.07}$ & $2.12_{\pm 0.08}$ & 0.89 \\
        & APS & 100 & 0.05 & 0.01 & 100.0 & $1.75_{\pm 0.05}$ & $2.08_{\pm 0.06}$ & 0.89 \\
        \bottomrule
    \end{tabular}
    }
\end{table*}
\begin{table*}[!h]
    \caption{Hyperparameter Search for EMNIST.}
    \centering
    \label{tab:hyper_emnist}
    \resizebox{\textwidth}{!}{  
    \begin{tabular}{c c | c c c c | c c c}
        Bound & Optimized for & batch size & lr & temperature & steepness & THR & APS & Test Acc.\\
        \midrule
        \multirow{ 2}{*}{CE} & THR & 100 & 0.01 & - & - & $2.06_{\pm 0.11}$ & $3.37_{\pm 0.15}$ & 0.86 \\
        & APS & 100 & 0.005 & - & - & $2.06_{\pm 0.10}$ & $3.40_{\pm 0.18}$ & 0.86 \\
        \cmidrule[0.5pt](l){1-9}
        \multirow{ 2}{*}{ConfTr} & THR & 100 & 0.01 & 0.1 & 100 & $1.99_{\pm 0.10}$ & $4.94_{\pm 0.26}$ & 0.85 \\
        & APS & 100 & 0.005 & 1.0 & 100 & $1.99_{\pm 0.08}$ & $2.36_{\pm 0.11}$ & 0.83\\
        \cmidrule[0.5pt](l){1-9}
        \multirow{ 2}{*}{ConfTr-class} & THR & 100 & 0.005 & 0.1 & 100 & $2.01_{\pm 0.09}$ & $5.07_{\pm 0.24}$ & 0.85 \\
        & APS & 100 & 0.05 & 1.0 & 100 & $1.99_{\pm 0.10}$ & $2.38_{\pm 0.11}$ & 0.84\\
        \cmidrule[0.5pt](l){1-9}
        \multirow{ 2}{*}{Fano} & THR & 100 & 0.01 & 0.1 & 100 & $2.10_{\pm 0.11}$ & $8.27_{\pm 0.53}$ & 0.84 \\
        & APS & 100 & 0.01 & 1.0 & 100 & $2.05_{\pm 0.11}$ & $2.75_{\pm 0.14}$ & 0.82\\
        \cmidrule[0.5pt](l){1-9}
        \multirow{ 2}{*}{MB Fano} & THR & 100 & 0.01 & 1.0 & 100 & $2.01_{\pm 0.11}$ & $5.21_{\pm 0.26}$ & 0.86\\
        & APS & 100 & 0.01 & 0.1 & 1 & $2.94_{\pm 0.17}$ & $3.67_{\pm 0.13}$ & 0.86\\
        \cmidrule[0.5pt](l){1-9}
        \multirow{ 2}{*}{DPI } & THR & 100 & 0.05 & 0.5 & 100 & $1.98_{\pm 0.09}$ & $3.86_{\pm 0.20}$ & 0.86 \\
        & APS & 100 & 0.05 & 0.1 & 100 & $2.04_{\pm 0.12}$ & $4.07_{\pm 0.23}$ & 0.86\\
        \bottomrule
    \end{tabular}
    }
\end{table*}
\begin{table*}[!h]
    \caption{Hyperparameter Search for CIFAR10.}
    \centering
    \label{tab:hyper_cifar10}
    \resizebox{\textwidth}{!}{
    \begin{tabular}{c c | c c c c | c c | c}
        Bound & Optimized for & batch size & lr & temperature & steepness & THR & APS & Test Acc.\\
        \midrule
        \multirow{ 2}{*}{CE} & THR & 100 & 0.05 & - & - & $1.69_{\pm 0.11}$ & $2.12_{\pm 0.21}$ & 0.93 \\
        & APS & 100 & 0.05 & - & - & $1.74_{\pm 0.07}$ & $2.34_{\pm 0.22}$ & 0.93 \\
        \cmidrule[0.5pt](l){1-9}
        \multirow{ 2}{*}{ConfTr} & THR & 100 & 0.05 & 0.5 & 10 & $9.90_{\pm 0.02}$ & $10.00_{\pm 0.00}$ & 0.10 \\
        & APS & 1000 & 0.005 & 0.1 & 1 & $9.90_{\pm 0.01}$ & $9.98_{\pm 0.00}$ & 0.10 \\
        \cmidrule[0.5pt](l){1-9}
        \multirow{ 2}{*}{ConfTr-class} & THR & 100 & 0.01 & 0.5 & 10 & $2.16_{\pm 0.09}$ & $2.19_{\pm 0.10}$ & 0.86 \\
        & APS & 100 & 0.01 & 0.5 & 10 & $2.13_{\pm 0.08}$ & $2.18_{\pm 0.06}$ & 0.86 \\
        \cmidrule[0.5pt](l){1-9}
        \multirow{ 2}{*}{Fano} & THR & 100 & 0.01 & 1.0 & 1 & $2.05_{\pm 0.05}$ & $2.34_{\pm 0.09}$ & 0.89 \\
        & APS & 100 & 0.01 & 1.0 & 1 & $2.06_{\pm 0.10}$ & $2.35_{\pm 0.10}$ & 0.89 \\
        \cmidrule[0.5pt](l){1-9}
        \multirow{ 2}{*}{MB Fano} & THR & 100 & 0.05 & 0.5 & 100 & $1.66_{\pm 0.09}$ & $2.40_{\pm 0.08}$ & 0.92 \\
        & APS & 100 & 0.01 & 1.0 & 10 & $1.69_{\pm 0.09}$ & $1.89_{\pm 0.06}$ & 0.91 \\
        \cmidrule[0.5pt](l){1-9}
        \multirow{ 2}{*}{DPI } & THR & 100 & 0.05 & 0.01 & 100 & $1.64_{\pm 0.07}$ & $1.88_{\pm 0.05}$ & 0.92 \\
        & APS & 100 & 0.005 & 0.01 & 10 & $1.79_{\pm 0.12}$ & $1.97_{\pm 0.08}$ & 0.91 \\
        \bottomrule
    \end{tabular}
    }
\end{table*}

\begin{table*}[!ht]
    \caption{Hyperparameter Search for CIFAR100.}
    \centering
    \label{tab:hyper_cifar100}
    \resizebox{\textwidth}{!}{
    \begin{tabular}{c c | c c c c | c c | c}
        Bound & Optimized for & batch size & lr & temperature & steepness & THR & APS & Test Acc.\\
        \midrule
        \multirow{ 2}{*}{CE} & THR & 100 & 0.05 & - & - & $19.70_{\pm 2.05}$ & $26.02_{\pm 1.31}$ & 0.72 \\
        & APS & 100 & 0.05 & - & - & $19.70_{\pm 2.05}$ & $26.02_{\pm 1.31}$ & 0.72 \\
        \cmidrule[0.5pt](l){1-9}
        \multirow{ 2}{*}{ConfTr} & THR & 100 & 0.005 & 1.0 & 1 & $32.80_{\pm 2.75}$ & $34.09_{\pm 2.54}$ &  0.52 \\
        & APS & 100 & 0.01 & 0.5 & 1 & $30.04_{\pm 1.36}$ & $40.58_{\pm 1.23}$ & 0.53 \\
        \cmidrule[0.5pt](l){1-9}
        \multirow{ 2}{*}{ConfTr-class} & THR & 100 & 0.01 & 1.0 & 1 & $66.48_{\pm 3.67}$ & $54.30_{\pm 17.12}$ & 0.43  \\
        & APS & 100 & 0.01 & 1 & 10 & $33.32_{\pm 1.89}$ & $32.91_{\pm 1.53}$ & 0.37  \\
        \cmidrule[0.5pt](l){1-9}
        \multirow{ 2}{*}{Fano} & THR & 100 & 0.05 & 0.5 & 1 & $40.30_{\pm 1.10}$ & $48.21_{\pm 1.26}$ & 0.42 \\
        & APS & 500 & 0.05 & 1 & 1 & $30.43_{\pm 1.61}$ & $33.80_{\pm 0.93}$ & 0.57 \\
        \cmidrule[0.5pt](l){1-9}
        \multirow{ 2}{*}{MB Fano} & THR & 100 & 0.05 & 0.5 & 100 & $14.61_{\pm 0.84}$ & $21.69_{\pm 0.71}$ & 0.70  \\
        & APS & 100 & 0.05 & 1 & 10 & $16.36_{\pm 0.93}$ & $21.68_{\pm 1.44}$ & 0.68 \\
        \cmidrule[0.5pt](l){1-9}
        \multirow{ 2}{*}{DPI } & THR & 100 & 0.05 & 1.0 & 1 & $17.55_{\pm 1.33}$ & $20.13_{\pm 0.78} $ & 0.69  \\
        & APS & 100 & 0.05 & 1 & 10 & $14.90_{\pm 0.80}$ & $17.41_{\pm 0.62}$ & 0.70  \\
        \bottomrule
    \end{tabular}
    }
\end{table*}

\vspace*{\fill}



\subsubsection{Results with Regularized Adaptive Prediction Sets (RAPS)} \label{app:raps}

We report additional results with a conformal method known as RAPS (regularized adaptive prediction sets) \citep{angelopoulos_uncertainty_2020}. In a nutshell, RAPS works exactly like APS but adds a penalization term $\lambda_{\text{reg}}$ to the score of each label that would make the prediction set larger than $k_{\text{reg}}.$
In Table~\ref{tab:central_inef_results_raps}, we have results with RAPS hyperparameters set as $k_{\text{reg}}{=}1$ and $\lambda_{\text{reg}}{=}0.01$, where we can see a pattern of performance across conformal training objectives that is similar to what we previously observed for THR and APS. As expected, the regularization introduced by RAPS results in prediction sets that are smaller than those obtained via APS in almost all cases. 
Again, similarly to \citep{stutz2022learning}, in all of our experiments we perform conformal training by thresholding on log-probabilities, which we observed to work best. The results we report for thresholding on probabilities (THR), APS and now RAPS show that this translates well to other score functions as well.
Nonetheless, the improvement we get via RAPS still does not offset the gains from conformal training. In most cases, RAPS applied to a model trained via CE produces less efficient prediction sets than APS applied to a model trained to optimize one of our bounds.

\input{tables/raps}

We also study the impact of the two hyperparameter in RAPS, namely $\lambda_{reg}$ and $k_{reg}$ in the inefficiency of each of the conformal training methods. We concentrate our analysis on CIFAR100 and report results with varying $k_{\text{reg}}$ in Table~\ref{tab:vary_kreg} and varying $\lambda_{\text{reg}}$ in Table~\ref{tab:vary_lamreg}. Interestingly, RAPS was much more sensitive to variations in $k_{\text{reg}}$ than in $\lambda_{\text{reg}}$. Still, we observe models trained via our model-based Fano and DPI bounds produce more efficient prediction sets than the baselines in all cases. 

\input{tables/vary_raps}

\subsection{Federated Setting}
In the federated setting, we run conformal training exactly in the same fashion, but including the additional $Q_{Z|X}$ term in (\ref{eq:fed_bounds}) to get the proper distributed bound that can be optimized locally and independently in each device. We optimize the conformal training objective with SGD for one epoch in each device, and then communicate the resulting ``personalized'' model to the server, which aggregates the model parameters of each device via federated averaging \cite{mcmahan2017communication}. After aggregation, the global model thus computed is communicated to the client and the process restarts. We do 5K such communication rounds for EMNIST, and 10K for CIFAR10 and CIFAR100. As described in the main text, we divide the data among 100, 500, and 1K clients for CIFAR10, CIFAR100 and EMNIST, respectively. We assign data points to devices imposing a \emph{distribution-based label imbalance} \cite{li2022federated}, where we sample a marginal label distribution for each device from a Dirichlet distribution. We use $\text{Dir}(1.0)$ for all experiments, but also study the effect of $\text{Dir}(0.5)$ and $\text{Dir}(0.1)$ on CIFAR10, as shown in Tables~\ref{tab:fed_thr_dir} and~\ref{tab:fed_aps_dir}. The remainder of the experimental setup is similar to that used for the centralized setting, with the same architectures, data augmentation strategies, and optimal hyperparameters reported in Tables~\ref{tab:hyper_mnist}, \ref{tab:hyper_fashion}, \ref{tab:hyper_emnist}, \ref{tab:hyper_cifar10}. 
In Tables~\ref{tab:fed_thr_dir} and~\ref{tab:fed_aps_dir}, as well as \ref{tab:per_glo_thr} and~\ref{tab:per_glo_aps}, we report inefficiency results for two different settings, global (GLO) and personalized (PER).

\paragraph{GLO} We run SCP with the final global model assuming calibration and test data sets at the server, or equivalently that the clients share their scores with the server. This reflects the best inefficiency results we can hope for with the global model, as in practice we might need to resort to privacy-preserving methods that are likely to hurt performance. Notably, we considered the quantile of quantiles approach of \citet{humbert2023one} as well as the simpler alternative proposed in \cite{lu2021distribution}, but in both cases we got varying degrees of coverage due to the data heterogeneity among devices introduced in the \emph{distribution-based label imbalance} setup. Addressing these shortcomings is a promising avenue for future research for conformal prediction in the federated setting.

\paragraph{PER} After learning the global model, we fine-tune it on the local training data of each device to obtain a personalized model. We then run SCP individually for each device with local calibration and test data sets and report the average prediction sets across all clients. Importantly, since each client has access to only a small number of data points, we do not always achieve valid coverage in the personalized setting. More precisely, all personalized models on CIFAR10 and EMNIST achieved marginal coverage of around 97\%, while for CIFAR100 that value dropped to 90\%. Nonetheless, all methods get similar coverages so the results remain comparable. Interestingly, even with personalized models, which implicitly already estimate $Q_{Y|X,Z}$, updating the personalized model as in (\ref{eq:model_with_si}) with the global head $Q_{Z|X,Y}$, still results in non-negligible improvements in performance, as shown in Tables~\ref{tab:fed_thr_dir} and~\ref{tab:fed_aps_dir}, as well as \ref{tab:per_glo_thr} and~\ref{tab:per_glo_aps}.

\input{tables/dirichlet_thr}
\input{tables/dirichlet_aps}

\input{tables/per_glo_thr}
\input{tables/per_glo_aps}
\vspace*{\fill}

\subsection{Evaluating the lower bounds on the set size} \label{app:set_size_estimate}
In this section, we evaluate our two lower bounds on the expected $[\log|C(X)|]^+$. The first one can be obtained by rearranging the simple Fano bound (c.f. Proposition~\ref{prop_app:fano_cp_hyx}) whereas the second one can be obtained from the model-based Fano entropy upper bounds (c.f. Theorem~\ref{prop:coverage_set_size_model_based_fano}). The main challenge in evaluating these bounds is in that we require the ground truth entropy $H(Y|X)$ or $H(Y|\hat{Y})$ which in general are not available. To proceed, we adopt the versions of the bounds that depend on $H(Y|\hat{Y})$. When $\hat{Y}$ is discrete, we can get a tractable lower bound to $H(Y|\hat{Y})$ via a maximum likelihood estimate of the entropy~\cite{paninski2003estimation}. More specifically, we have that
\begin{align*}
H(Y|\hat{Y}) = H(Y, \hat{Y}) - H(\hat{Y}) \geq H_{MLE}(Y, \hat{Y}) - \log |\hat{\mathcal{Y}}|,
\end{align*}
where $|\hat{\mathcal{Y}}|$ is the cardinality of $\hat{Y}$.

\begin{figure}[htb]
    \centering
    \begin{subfigure}[t]{0.49\textwidth}
         \centering
         \includegraphics[width=\textwidth]{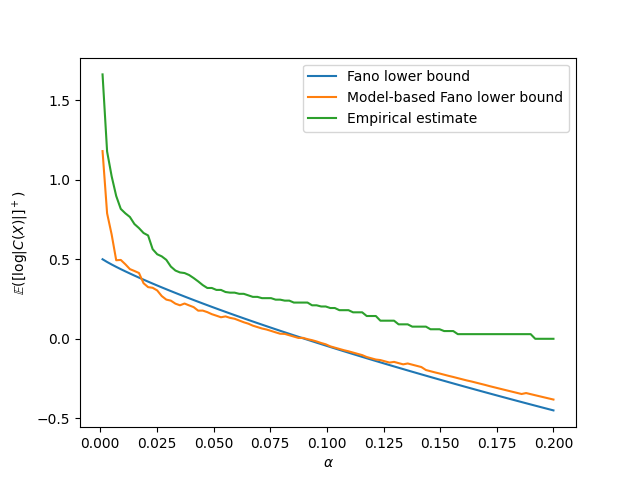}
        \caption{CIFAR 10}
     \end{subfigure}
     \begin{subfigure}[t]{0.49\textwidth}
         \centering
         \includegraphics[width=\textwidth]{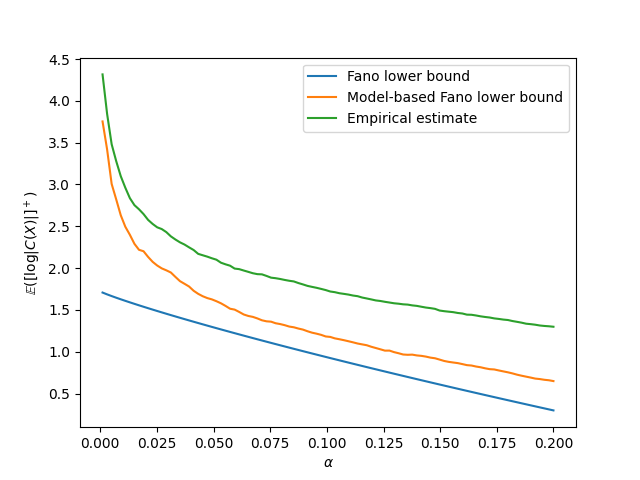}
        \caption{CIFAR 100}
     \end{subfigure}
    \caption{Expected $[\log|C(X)|]^+$ as a function of $\alpha$.}
    \label{fig:set_size_exps}
\end{figure}

Based on this, we evaluate our set size bounds on ResNet models trained with CE on CIFAR 10 and CIFAR 100. As the logits $\hat{Y}$ used for CP are not discrete, we perform K-means clustering on them and construct a vector quantized $\hat{Y}_{vq}$ by assigning to each logit vector its closest cluster centroid. We then use $\hat{Y}_{vq}$ to perform CP with thresholding and also use $\hat{Y}_{vq}$ to obtain a lower bound on $H(Y|\hat{Y}_{vq})$ via a maximum likelihood estimate for $H(Y,\hat{Y}_{vq})$. For this maximum likelihood estimate we also perform the Miller-Madow bias correction~\cite{paninski2003estimation}. 

For CIFAR 10 we cluster the logits into 32 clusters whereas for CIFAR 100 we use 256 clusters. The centroids are learned on a calibration set of 5k logits. For the model-based Fano bound, as it needs to compute terms that depend on the model probabilities and on whether the label was correctly covered by CP or not, we further split the calibration set into two equal-sized chunks; the first is used to find the quantile for thresholding and the second one is used to evaluate the terms of the bound. The obtained lower bounds on $[\log|C(X)|]^+$ for various $\alpha$'s can be seen at Figure~\ref{fig:set_size_exps}. We also include an ``empirical estimate'' which is obtained by computing the quantile with the quantized calibration logits and then measuring the average $[\log|C(X)|]^+$ on the test set via thresholding on the quantized test logits. We see that model-based Fano provides relatively tight estimates for small values of alpha.

%% file: tables/raps.tex
\begin{table*}[!ht]
\centering
\caption{\textbf{Inefficiency restuls with RAPS}. Average prediction set size with both APS and RAPS ($k_{\text{reg}}=1$ and $\lambda_{\text{reg}} = 0.01$) methods for each conformal training objective and data set for a target $\alpha = 0.01$. We report mean and standard deviation across 10 different calib./test splits and show in bold all values within one standard deviation of the best result. Both APS and RAPS results were computed with the same classifier. Lower is better.}
\label{tab:central_inef_results_raps}
\resizebox{\textwidth}{!}{  
\begin{tabular}{l@{\hskip 8pt}  c@{\hskip 4pt}c@{\hskip 8pt}  c@{\hskip 4pt}c@{\hskip 8pt}  c@{\hskip 4pt}c@{\hskip 8pt}  c@{\hskip 4pt}c@{\hskip 8pt}  c@{\hskip 4pt}c}
\toprule
 Method & \multicolumn{2}{c}{MNIST} & \multicolumn{2}{c}{F-MNIST} & \multicolumn{2}{c}{EMNIST} & \multicolumn{2}{c}{CIFAR 10} & \multicolumn{2}{c}{CIFAR 100}\\\midrule
 & APS & RAPS & APS & RAPS & APS & RAPS & APS & RAPS & APS & RAPS\\\midrule
CE & $2.50_{\pm0.08}$ & $2.35_{\pm0.16}$ & $2.41_{\pm0.17}$ & $2.40_{\pm0.17}$ & $3.40_{\pm0.18}$ & $3.07_{\pm0.16}$ & $2.34_{\pm0.22}$ & $2.31_{\pm0.22}$& $26.02_{\pm1.31}$ & $26.02_{\pm1.31}$\\\midrule
ConfTr & $\mathbf{2.10_{\pm0.07}}$ & $\mathbf{2.10_{\pm0.07}}$ & $\mathbf{1.89_{\pm0.09}}$& $1.84_{\pm0.09}$& $\mathbf{2.36_{\pm0.11}}$ & $\mathbf{2.19_{\pm0.06}}$ & $9.99_{\pm0.00}$ & $9.99_{\pm0.00}$ & $40.59_{\pm1.23}$& $39.87_{\pm1.02}$\\
ConfTr$_{\text{class}}$ & $\mathbf{2.13_{\pm0.13}}$ & $\mathbf{2.13_{\pm 0.13}}$ & $\mathbf{1.79_{\pm0.09}}$& $\mathbf{1.79_{\pm0.09}}$ & $\mathbf{2.38_{\pm0.11}}$ & $\mathbf{2.21_{\pm0.07}}$ & $2.18_{\pm0.06}$ & $2.18_{\pm0.06}$ & $32.91_{\pm1.53}$ & $32.94_{\pm1.54}$ \\\midrule
Fano & $\mathbf{2.12_{\pm0.08}}$ & $\mathbf{2.12_{\pm0.08}}$ & $\mathbf{1.87_{\pm0.05}}$& $1.80_{\pm0.04}$ & $\mathbf{2.75_{\pm0.14}}$ & $2.50_{\pm0.07}$ & $2.35_{\pm0.10}$ & $2.20_{\pm0.04}$ & $33.80_{\pm0.93}$& $33.75_{\pm1.02}$\\
MB Fano & $\mathbf{2.49_{\pm0.19}}$ & $\mathbf{2.48_{\pm0.19}}$ & $\mathbf{2.25_{\pm0.14}}$ & $2.07_{\pm0.11}$& $3.67_{\pm0.13}$ & $3.67_{\pm0.13}$ & $\mathbf{1.89_{\pm0.06}}$ & $\mathbf{1.78_{\pm0.07}}$ & $\mathbf{21.68_{\pm1.44}}$ & $20.97_{\pm1.30}$ \\
DPI & $\mathbf{2.64_{\pm0.07}}$ & $2.44_{\pm0.12}$ & $\mathbf{2.08_{\pm0.06}}$ & $1.95_{\pm0.05}$ & $\mathbf{4.07_{\pm0.23}}$ & $3.74_{\pm0.17}$ & $\mathbf{1.97_{\pm0.08}}$ & $\mathbf{1.87_{\pm0.11}}$ & $\mathbf{17.41_{\pm0.62}}$ & $\mathbf{17.25_{\pm0.58}}$ \\
\bottomrule
\end{tabular}
}
\end{table*}

%% file: tables/vary_raps.tex
\begin{table*}[!ht]
\centering
\caption{\textbf{Inefficiency results for RAPS with varying $k_{\text{reg}}$}. Average prediction set size
for $\lambda_{\text{reg}} = 0.01$ and varying $k_{\text{reg}}$ on CIFAR100 for a target $\alpha = 0.01$. We report mean and standard deviation across 10 different calib./test splits and show in bold all values within one standard deviation of the best result. Both APS and RAPS results were computed with the same classifier. Lower is better.}
\label{tab:vary_kreg}
\resizebox{\textwidth}{!}{  
\begin{tabular}{l  cc  cc  cc  cc}
\toprule
 Method & $k_{\text{reg}}=0$ (APS) & $k_{\text{reg}}=1$ & $k_{\text{reg}}=3$ & $k_{\text{reg}}=5$ & $k_{\text{reg}}=10$ & $k_{\text{reg}}=15$\\\midrule
CE & $26.02_{\pm1.31}$ & $26.02_{\pm1.31}$ & $22.43_{\pm1.88}$ & $20.43_{\pm1.70}$ & $18.95_{\pm1.18}$ & $20.63_{\pm0.89}$\\\midrule
ConfTr & $40.59_{\pm1.23}$ & $39.87_{\pm1.02}$ & $38.73_{\pm0.84}$ & $38.31_{\pm0.70}$ & $38.06_{\pm0.48}$ & $38.06_{\pm0.59}$\\
ConfTr$_{\text{class}}$ & $32.91_{\pm1.53}$ & $32.94_{\pm1.54}$ & $32.94_{\pm1.53}$ & $32.95_{\pm 1.53}$ & $33.16_{\pm1.47}$ & $33.74_{\pm1.39}$ \\\midrule
Fano & $33.80_{\pm0.93}$& $33.75_{\pm1.02}$ & $32.83_{\pm1.60}$ & $31.71_{\pm1.55}$ & $30.82_{\pm1.72}$ & $30.31_{\pm1.71}$\\
MB Fano & $21.68_{\pm1.44}$ & $20.97_{\pm1.30}$ & $19.32_{\pm1.40}$ & $19.04_{\pm1.36}$ & $\mathbf{17.35_{\pm0.86}}$ & $\mathbf{17.68_{\pm0.94}}$\\
DPI & $\mathbf{17.41_{\pm0.62}}$ & $\mathbf{17.25_{\pm0.58}}$ & $\mathbf{17.02_{\pm0.75}}$ & $\mathbf{16.40_{\pm0.93}}$ & $\mathbf{17.68_{\pm0.82}}$ & $19.49_{\pm2.08}$\\
\bottomrule
\end{tabular}
}
\end{table*}

\begin{table*}[!ht]
\centering
\caption{\textbf{Inefficiency results for RAPS with varying $\lambda_{\text{reg}}$}. Average prediction set size
for $k_{\text{reg}} = 1$ and varying $\lambda_{\text{reg}}$ on CIFAR100 for a target $\alpha = 0.01$. We report mean and standard deviation across 10 different calib./test splits and show in bold all values within one standard deviation of the best result. Both APS and RAPS results were computed with the same classifier. Lower is better.}
\label{tab:vary_lamreg}
\resizebox{\textwidth}{!}{  
\begin{tabular}{l  cc  cc  cc  cc}
\toprule
 Method & $\lambda_{\text{reg}}=0$ (APS) & $\lambda_{\text{reg}}=0.01$ & $\lambda_{\text{reg}}=0.03$ & $\lambda_{\text{reg}}=0.1$ & $\lambda_{\text{reg}}=0.3$ & $\lambda_{\text{reg}}=1.0$\\\midrule
CE & $26.02_{\pm1.31}$ & $26.02_{\pm1.31}$ & $26.02_{\pm1.31}$ & $26.02_{\pm1.31}$ & $26.02_{\pm1.31}$ & $26.02_{\pm1.31}$\\\midrule
ConfTr & $40.59_{\pm1.23}$ & $39.87_{\pm1.02}$ & $39.17_{\pm1.05}$ & $39.09_{\pm1.12}$ & $39.09_{\pm1.12}$ & $39.09_{\pm1.12}$\\
ConfTr$_{\text{class}}$ & $32.91_{\pm1.53}$ & $32.94_{\pm1.54}$ & $32.94_{\pm1.54}$ & $32.93_{\pm 1.53}$ & $32.94_{\pm 1.53}$ & $32.94_{\pm1.55}$ \\\midrule
Fano & $33.80_{\pm0.93}$& $33.75_{\pm1.02}$ & $33.75_{\pm1.02}$ & $33.75_{\pm1.02}$ & $33.75_{\pm1.02}$ & $33.75_{\pm1.02}$\\
MB Fano & $21.68_{\pm1.44}$ & $20.97_{\pm1.30}$ & $20.96_{\pm1.29}$ & $20.96_{\pm1.29}$ & $20.96_{\pm1.29}$ & $20.96_{\pm1.29}$\\
DPI & $\mathbf{17.41_{\pm0.62}}$ & $\mathbf{17.25_{\pm0.58}}$ & $\mathbf{17.25_{\pm0.58}}$ & $\mathbf{17.25_{\pm0.58}}$ & $\mathbf{17.25_{\pm0.58}}$ & $\mathbf{17.25_{\pm0.58}}$\\
\bottomrule
\end{tabular}
}
\end{table*}

%% file: tables/dirichlet_thr.tex
\begin{table*}[!htb]
\centering
\caption{\textbf{\textbf{Inefficiency results for different degrees of data heterogeneity in FL with THR}}. Average prediction set size with THR for CIFAR10 with different levels of data heterogeneity among clients for both global (GLO) and personalized (PER) models in the federated setting for a target $\alpha = 0.01$. We use $_{+\text{SI}}$ to indicate the inclusion of side information. Lower is better.}
\label{tab:fed_thr_dir}
\resizebox{\textwidth}{!}{  
\begin{tabular}{ l@{\hskip 6pt} c@{\hskip 3pt}c@{\hskip 3pt}c@{\hskip 3pt}c@{\hskip 6pt}  c@{\hskip 3pt}c@{\hskip 3pt}c@{\hskip 3pt}c@{\hskip 6pt} c@{\hskip 3pt}c@{\hskip 3pt}c@{\hskip 3pt}c@{\hskip 6pt} }
\toprule
 Method & \multicolumn{4}{c}{Dirichlet(1.0)} & \multicolumn{4}{c}{Dirichlet(0.5)} & \multicolumn{4}{c}{Dirichlet(0.1)} \\\midrule
 & GLO & GLO$_{+\text{SI}}$ & PER & PER$_{+\text{SI}}$ & GLO & GLO$_{+\text{SI}}$ & PER & PER$_{+\text{SI}}$ & GLO & GLO$_{+\text{SI}}$ & PER & PER$_{+\text{SI}}$ \\\midrule
CE & $2.73_{\pm 0.04}$  & $2.30_{\pm 0.06}$  & $2.07_{\pm 0.51} $  & $1.82_{\pm 0.39}$ & $2.81_{\pm 0.11}$  & $2.41_{\pm 0.08}$ & $2.03_{\pm 0.49}$  & $1.79_{\pm 0.38}$ & $2.73_{\pm 0.13}$  & $2.32_{\pm 0.06}$ & $2.05_{\pm 0.52}$  & $1.82_{\pm 0.38}$  \\
\midrule
ConfTr & $10.00_{\pm 0.00}$  & $10.00_{\pm 0.00}$  & $9.84_{\pm 0.05}$  & $9.84_{\pm 0.05}$ & $10.00_{\pm 0.00}$  & $10.00_{\pm 0.00}$ & $9.84_{\pm 0.05}$  & $9.84_{\pm 0.05}$ & $10.00_{\pm 0.00}$  & $10.00_{\pm 0.00}$ & $9.87_{\pm 0.06}$  & $9.87_{\pm 0.06}$  \\
ConfTr$_{\text{class}}$ & $ 3.53_{\pm 0.09}$  & $3.39_{\pm 0.08}$  & $2.86_{\pm 0.38} $  & $2.79_{\pm 0.35}$ & $3.54_{\pm 0.06}$  & $3.38_{\pm 0.07}$ & $2.91_{\pm 0.40}$  & $2.84_{\pm 0.37}$ & $3.53_{\pm 0.05}$  & $3.39_{\pm 0.05}$ & $2.84_{\pm 0.41}$  & $2.78_{\pm 0.38}$ \\
\midrule
Fano & $2.39_{\pm 0.07}$  & $2.07_{\pm 0.07}$  & $1.84_{\pm 0.37} $  & $1.63_{\pm 0.30}$ & $2.47_{\pm 0.08}$  & $2.10_{\pm 0.06}$ & $1.94_{\pm 0.43}$  & $1.71_{\pm 0.35}$ & $2.46_{\pm 0.08}$  & $2.11_{\pm 0.06}$ & $1.91_{\pm 0.41}$  & $1.68_{\pm 0.33}$  \\
MB Fano & $2.52_{\pm 0.08}$  & $2.04_{\pm 0.07}$  & $1.56_{\pm 0.29} $  & $1.40_{\pm 0.22}$ & $2.66_{\pm 0.12}$  & $2.09_{\pm 0.06}$ & $1.57_{\pm 0.31}$  & $1.40_{\pm 0.23}$ & $2.55_{\pm 0.11}$  & $2.06_{\pm 0.10}$ & $1.98_{\pm 0.45}$  & $1.64_{\pm 0.32}$ \\
DPI & $2.76_{\pm 0.07}$  & $2.28_{\pm 0.03}$  & $1.64_{\pm 0.36} $  & $1.49_{\pm 0.28}$ & $2.50_{\pm 0.07}$  & $2.11_{\pm 0.49}$ & $1.91_{\pm 0.42}$  & $1.64_{\pm 0.31}$ & $2.51_{\pm 0.07}$  & $2.07_{\pm 0.05}$ & $2.04_{\pm 0.41}$  & $1.70_{\pm 0.30}$  \\
\bottomrule
\end{tabular}
}
\end{table*}

%% file: tables/dirichlet_aps.tex
\begin{table*}[!h]
\centering
\caption{\textbf{Inefficiency results for different degrees of data heterogeneity in FL with APS}. Average prediction set size with APS for CIFAR10 with different levels of data heterogeneity among clients for both global (GLO) and personalized (PER) models in the federated setting for a target $\alpha = 0.01$. We use $_{+\text{SI}}$ to indicate the inclusion of side information. Lower is better.}
\label{tab:fed_aps_dir}
\resizebox{\textwidth}{!}{  
\begin{tabular}{ l@{\hskip 6pt} c@{\hskip 3pt}c@{\hskip 3pt}c@{\hskip 3pt}c@{\hskip 6pt}  c@{\hskip 3pt}c@{\hskip 3pt}c@{\hskip 3pt}c@{\hskip 6pt} c@{\hskip 3pt}c@{\hskip 3pt}c@{\hskip 3pt}c@{\hskip 6pt} }
\toprule
 Method & \multicolumn{4}{c}{Dirichlet(1.0)} & \multicolumn{4}{c}{Dirichlet(0.5)} & \multicolumn{4}{c}{Dirichlet(0.1)} \\\midrule
 & GLO & GLO$_{+\text{SI}}$ & PER & PER$_{+\text{SI}}$ & GLO & GLO$_{+\text{SI}}$ & PER & PER$_{+\text{SI}}$ & GLO & GLO$_{+\text{SI}}$ & PER & PER$_{+\text{SI}}$ \\\midrule
CE & $2.83_{\pm 0.07}$  & $2.43_{\pm 0.06}$   & $2.22_{\pm 0.06}$   & $1.94_{\pm 0.41}$  & $2.70_{\pm 0.79}$   & $2.33_{\pm 0.42}$  & $ 2.13_{\pm 0.47}$   & $1.87_{\pm 0.36}$  & $ 2.81_{\pm 0.14}$   & $2.37_{\pm 0.08}$  & $2.15_{\pm 0.05}$   & $1.89_{\pm 0.39}$   \\
\midrule
ConfTr & $10.00_{\pm 0.00}$  & $10.00_{\pm 0.00}$  & $9.87_{\pm 0.14}$   & $9.87_{\pm 0.14}$  & $10.00_{\pm 0.00}$  & $10.00_{\pm 0.00}$  & $9.88_{\pm  0.09}$   & $9.88_{\pm  0.09}$  & $10.00_{\pm 0.00}$  & $10.00_{\pm 0.00}$  & $9.87_{\pm 0.14}$   & $9.87_{\pm 0.14}$  \\
ConfTr$_{\text{class}}$ & $10.00_{\pm 0.00}$  & $10.00_{\pm 0.00}$   & $9.92_{\pm 0.03}$  & $9.92_{\pm 0.03}$  & $10.00_{\pm 0.00}$  & $10.00_{\pm 0.00}$  & $ 9.94_{\pm 0.07}$   & $9.95_{\pm 0.05}$  & $10.00_{\pm 0.00}$  & $10.00_{\pm 0.00}$  & $9.92_{\pm 0.03}$   & $9.92_{\pm 0.03}$ \\
\midrule
Fano & $2.73_{\pm 0.07}$  & $2.39_{\pm 0.06}$   & $2.17_{\pm 0.45}$   & $1.94_{\pm 0.36}$  & $2.61_{\pm 0.07}$   & $2.30_{\pm 0.05}$  & $ 2.15_{\pm 0.04}$   & $1.92_{\pm 0.34}$  & $ 2.67_{\pm 0.08}$   & $2.35_{\pm 0.07}$  & $2.13_{\pm 0.42}$   & $1.92_{\pm 0.34}$  \\
MB Fano & $2.79_{\pm 0.13}$  & $2.33_{\pm 0.05}$   & $2.28_{\pm 0.48}$   & $1.97_{\pm 0.35}$  & $2.71_{\pm 0.09}$   & $2.31_{\pm 0.07}$  & $2.27_{\pm 0.48}$   & $ 1.96_{\pm 0.35}$  & $2.87_{\pm 0.12}$   & $2.40_{\pm 0.06}$  & $2.32_{\pm 0.51}$   & $1.99_{\pm 0.39}$ \\
DPI & $2.68_{\pm 0.15}$  & $2.22_{\pm 0.09}$   & $2.11_{\pm 0.51}$   & $1.84_{\pm 0.36}$  & $2.54_{\pm 0.10}$   & $2.18_{\pm 0.04}$  & $2.10_{\pm 0.45}$   & $ 1.84_{\pm 0.33}$  & $ 2.65_{\pm 0.10}$   & $2.25_{\pm 0.08}$  & $2.13_{\pm 0.44}$   & $1.89_{\pm 0.34}$  \\
\bottomrule
\end{tabular}
}
\end{table*}

%% file: tables/per_glo_thr.tex
\begin{table*}[!ht]
\centering
\caption{\textbf{Inefficiency results for global (GLO) and personalized (PER) models with THR}. Average prediction set size with THR for federated conformal training with a target $\alpha = 0.01$. We use $_{+\text{SI}}$ to indicate the inclusion of side information. We show in bold all values within one standard deviation of the best result. Lower is better.}
\label{tab:per_glo_thr}
\resizebox{\textwidth}{!}{  
\begin{tabular}{ l@{\hskip 6pt} c@{\hskip 3pt}c@{\hskip 3pt}c@{\hskip 3pt}c@{\hskip 6pt} c@{\hskip 3pt}c@{\hskip 3pt}c@{\hskip 3pt}c@{\hskip 6pt}  c@{\hskip 3pt}c@{\hskip 3pt}c@{\hskip 3pt}c@{\hskip 6pt} }
\toprule
 Method & \multicolumn{4}{c}{EMNIST} & \multicolumn{4}{c}{CIFAR 10} & \multicolumn{4}{c}{CIFAR 100} \\\midrule
 & GLO & GLO$_{+\text{SI}}$ & PER & PER$_{+\text{SI}}$ & GLO & GLO$_{+\text{SI}}$ & PER & PER$_{+\text{SI}}$ & GLO & GLO$_{+\text{SI}}$ & PER & PER$_{+\text{SI}}$ \\\midrule
CE & $2.91_{\pm 0.02}$  & $2.46_{\pm 0.02}$ & $\mathbf{2.46_{\pm 0.65}}$  & $\mathbf{2.08_{\pm 0.51}}$ & $2.73_{\pm 0.04}$  & $2.30_{\pm 0.06}$  & $\mathbf{2.07_{\pm 0.51}}$  & $\mathbf{1.82_{\pm 0.39}}$  & $55.41_{\pm 1.09}$  & $52.31_{\pm 1.03}$  & $\mathbf{20.84_{\pm 8.02}}$  & $\mathbf{18.85_{\pm 7.53}}$  \\
\midrule
ConfTr & $4.60_{\pm 0.05}$  & $3.30_{\pm 0.02}$  & $\mathbf{3.95_{\pm 1.05}}$  & $3.05_{\pm 0.78}$ & $10.00_{\pm 0.00}$  & $10.00_{\pm 0.00}$  & $9.84_{\pm 0.05}$  & $9.84_{\pm 0.05}$  & $\mathbf{45.60_{\pm 1.30}}$  & $\mathbf{41.18_{\pm 1.16}}$  & $\mathbf{16.23_{\pm 6.27}}$  &  $\mathbf{14.01_{\pm 5.66}}$  \\
ConfTr$_{\text{class}}$ & $2.88_{\pm 0.02}$  & $\mathbf{1.98_{\pm 0.02}}$  & $\mathbf{1.69_{\pm 0.31}}$  & $\mathbf{1.55_{\pm 0.24}}$  & $3.53_{\pm 0.09}$  & $3.39_{\pm 0.08}$  & $2.86_{\pm 0.38} $  & $2.79_{\pm 0.35}$  & $58.53_{\pm 1.40}$  & $56.03_{\pm 1.29}$  & $\mathbf{25.22_{\pm 7.40}}$  &  $\mathbf{23.87_{\pm 7.02}}$  \\
\midrule
Fano & $\mathbf{2.63_{\pm 0.02}}$  & $2.37_{\pm 0.02}$  & $\mathbf{2.14_{\pm 0.46}}$  & $\mathbf{1.98_{\pm 0.40}}$  & $\mathbf{2.39_{\pm 0.07}}$  & $\mathbf{2.07_{\pm 0.07}}$  & $\mathbf{1.84_{\pm 0.37}}$  & $\mathbf{1.63_{\pm 0.30}}$  & $\mathbf{47.91_{\pm 1.20}}$  & $\mathbf{41.19_{\pm 1.02}}$  & $\mathbf{18.42_{\pm 6.95}}$  & $\mathbf{14.55_{\pm 5.83}}$ \\
MB Fano & $2.84_{\pm 0.04}$  & $2.25_{\pm 0.03}$ & $\mathbf{2.47_{\pm 0.64}}$ & $\mathbf{1.96_{\pm 0.45}}$  & $\mathbf{2.52_{\pm 0.08}}$  & $\mathbf{2.04_{\pm 0.07}}$  & $\mathbf{1.56_{\pm 0.29}}$  & $\mathbf{1.40_{\pm 0.22}}$  & $52.94_{\pm 1.40}$  & $46.97_{\pm 1.30}$ & $\mathbf{20.36_{\pm 7.77}}$ & $\mathbf{16.41_{\pm 6.79}}$   \\
DPI & $\mathbf{2.60_{\pm 0.02}}$  & $2.23_{\pm 0.01}$  & $\mathbf{2.31_{\pm 0.58}}$  & $\mathbf{1.97_{\pm 0.46}}$  & $2.76_{\pm 0.07}$  & $2.28_{\pm 0.03}$  & $\mathbf{1.64_{\pm 0.36}}$  & $\mathbf{1.49_{\pm 0.28}}$  & $52.36_{\pm 0.95}$  & $48.64_{\pm 0.70}$  & $\mathbf{20.13_{\pm 7.84}}$  & $\mathbf{17.73_{\pm 7.23}}$  \\
\bottomrule
\end{tabular}
}
\end{table*}

%% file: tables/per_glo_aps.tex
\begin{table*}[!ht]
\centering
\caption{\textbf{Inefficiency results for global (GLO) and personalized (PER) models with APS}. Average prediction set size with APS for federated conformal training with a target $\alpha = 0.01$. We use $_{+\text{SI}}$ to indicate the inclusion of side information. We show in bold all values within one standard deviation of the best result. Lower is better.}
\label{tab:per_glo_aps}
\resizebox{\textwidth}{!}{  
\begin{tabular}{ l@{\hskip 6pt}  c@{\hskip 3pt}c@{\hskip 3pt}c@{\hskip 3pt}c@{\hskip 6pt} c@{\hskip 3pt}c@{\hskip 3pt}c@{\hskip 3pt}c@{\hskip 6pt}  c@{\hskip 3pt}c@{\hskip 3pt}c@{\hskip 3pt}c@{\hskip 6pt} }
\toprule
 Method & \multicolumn{4}{c}{EMNIST} & \multicolumn{4}{c}{CIFAR 10} & \multicolumn{4}{c}{CIFAR 100} \\\midrule
 & GLO & GLO$_{+\text{SI}}$ & PER & PER$_{+\text{SI}}$ & GLO & GLO$_{+\text{SI}}$ & PER & PER$_{+\text{SI}}$ & GLO & GLO$_{+\text{SI}}$ & PER & PER$_{+\text{SI}}$ \\\midrule
CE & $3.69_{\pm 0.03}$ & $3.14_{\pm 0.04}$ & $1.42_{\pm 0.25}$  & $1.40_{\pm 0.24}$ & $\mathbf{2.83_{\pm 0.07}}$  & $2.43_{\pm 0.06}$  & $\mathbf{2.22_{\pm 0.56}}$  & $\mathbf{1.94_{\pm 0.41}}$  & $64.73_{\pm 0.34}$  & $62.67_{\pm 3.68}$  & $\mathbf{22.39_{\pm 8.86}}$  & $\mathbf{20.22_{\pm 8.15}}$  \\
\midrule
ConfTr & $6.14_{\pm 0.04}$  & $5.25_{\pm 0.04}$  & $4.79_{\pm 1.23}$  & $3.06_{\pm 0.63}$ & $10.00_{\pm 0.00}$  & $10.00_{\pm 0.00}$  & $9.87_{\pm 0.14}$  & $9.87_{\pm 0.14}$  & $55.18_{\pm 2.10}$  & $47.58_{\pm 1.48}$  & $\mathbf{23.43_{\pm 8.20}}$  &  $\mathbf{19.12_{\pm 7.14}}$  \\
ConfTr$_{\text{class}}$ & $\mathbf{2.65_{\pm 0.02}}$  & $\mathbf{2.42_{\pm 0.02}}$  & $3.03_{\pm 1.57}$  & $1.51_{\pm 0.31}$  & $10.00_{\pm 0.00}$  & $10.00_{\pm 0.00}$  & $9.92_{\pm 0.03}$  & $9.92_{\pm 0.03}$  & $99.92_{\pm 0.02}$  & $99.91_{\pm 0.01}$  & $99.97_{\pm 0.22}$ &  $99.68_{\pm 0.24}$  \\
\midrule
Fano & $3.12_{\pm 0.04}$  & $2.72_{\pm 0.03}$  & $\mathbf{1.17_{\pm 0.10}}$  & $\mathbf{1.15_{\pm 0.01}}$  & $\mathbf{2.73_{\pm 0.07}}$  & $\mathbf{2.39_{\pm 0.06}}$ & $\mathbf{2.17_{\pm 0.45}}$  & $\mathbf{1.94_{\pm 0.36}}$  & $\mathbf{46.95_{\pm 0.67}}$  & $\mathbf{42.75_{\pm 0.91}}$  & $\mathbf{19.19_{\pm 7.24}}$  & $\mathbf{16.9_{\pm 6.67}}$  \\
MB Fano & $4.75_{\pm 0.03}$  &  $\mathbf{2.43_{\pm 0.01}}$ & $2.04_{\pm 0.51}$ & $2.31_{\pm 0.42}$ & $\mathbf{2.79_{\pm 0.13}}$  & $\mathbf{2.33_{\pm 0.05}}$  & $\mathbf{2.28_{\pm 0.48}}$  & $\mathbf{1.97_{\pm 0.35}}$  & $50.72_{\pm 1.77}$  & $45.72_{\pm 1.38}$ & $\mathbf{21.12_{\pm 7.49}}$ & $\mathbf{18.12_{\pm 6.84}}$   \\
DPI & $2.98_{\pm 0.03}$  & $2.58_{\pm 0.02}$   & $2.73_{\pm 0.65}$  & $1.29_{\pm 0.16}$  & $\mathbf{2.68_{\pm 0.15}}$  & $\mathbf{2.22_{\pm 0.09}}$  & $\mathbf{2.11_{\pm 0.51}}$  & $\mathbf{1.84_{\pm 0.36}}$  & $51.29_{\pm 1.07}$  & $47.18_{\pm 1.27}$  & $\mathbf{20.76_{\pm 7.51}}$  & $\mathbf{18.15_{\pm 6.82}}$  \\
\bottomrule
\end{tabular}
}
\end{table*}

%% file: appendix/neurips_checklist.tex
\clearpage
\section*{NeurIPS Paper Checklist}

\begin{enumerate}

\item {\bf Claims}
    \item[] Question: Do the main claims made in the abstract and introduction accurately reflect the paper's contributions and scope?
    \item[] Answer: \answerYes{} 
    \item[] Justification: All of our claims are supported by our experimental results, and we provide detailed proofs for all theoretical results in the supplemental material.

\item {\bf Limitations}
    \item[] Question: Does the paper discuss the limitations of the work performed by the authors?
    \item[] Answer: \answerYes{} 
    \item[] Justification: We comment on the limitations of our work in Appendix~\ref{app:broader_impact}.

\item {\bf Theory Assumptions and Proofs}
    \item[] Question: For each theoretical result, does the paper provide the full set of assumptions and a complete (and correct) proof?
    \item[] Answer: \answerYes{} 
    \item[] Justification: We include detailed proofs for all theoretical results, clearly stating the assumptions (mainly exchangeable data and a valid conformal prediction method). We also provide the necessary background on information theory and conformal prediction in Appendix~\ref{app:background} to help the reader to follow our derivations.

    \item {\bf Experimental Result Reproducibility}
    \item[] Question: Does the paper fully disclose all the information needed to reproduce the main experimental results of the paper to the extent that it affects the main claims and/or conclusions of the paper (regardless of whether the code and data are provided or not)?
    \item[] Answer: \answerYes{} 
    \item[] Justification: We did our best to provide as many experimental details as possible, which can be found in the main text as well as in Appendix~\ref{app:experiments}.

\item {\bf Open access to data and code}
    \item[] Question: Does the paper provide open access to the data and code, with sufficient instructions to faithfully reproduce the main experimental results, as described in supplemental material?
    \item[] Answer: \answerYes{} 
    \item[] Justification: The source code will be hosted at \href{https://github.com/Qualcomm-AI-research/info_cp}{\nolinkurl{github.com/Qualcomm-AI-research/info_cp}}.

\item {\bf Experimental Setting/Details}
    \item[] Question: Does the paper specify all the training and test details (e.g., data splits, hyperparameters, how they were chosen, type of optimizer, etc.) necessary to understand the results?
    \item[] Answer: \answerYes{} 
    \item[] Justification: We give a detailed description of all of our experiments in Appendix~\ref{app:experiments}.

\item {\bf Experiment Statistical Significance}
    \item[] Question: Does the paper report error bars suitably and correctly defined or other appropriate information about the statistical significance of the experiments?
    \item[] Answer: \answerYes{} 
    \item[] Justification: We report one standard deviation confidence intervals, computed across 10 random calibration/test splits, for all of our experiments.

\item {\bf Experiments Compute Resources}
    \item[] Question: For each experiment, does the paper provide sufficient information on the computer resources (type of compute workers, memory, time of execution) needed to reproduce the experiments?
    \item[] Answer: \answerYes{} 
    \item[] Justification: Our experiments are not particularly demanding in the terms of compute and can be reproduced on any modern hardware. We do state in Appendix~\ref{app:experiments} that we ran our experiments in commercially available NVIDIA GPUs.
    
\item {\bf Code Of Ethics}
    \item[] Question: Does the research conducted in the paper conform, in every respect, with the NeurIPS Code of Ethics \url{https://neurips.cc/public/EthicsGuidelines}?
    \item[] Answer: \answerYes{} 
    \item[] Justification: Our work is in accordance with the NeurIPS Code of Ethics.

\item {\bf Broader Impacts}
    \item[] Question: Does the paper discuss both potential positive societal impacts and negative societal impacts of the work performed?
    \item[] Answer: \answerYes{} 
    \item[] Justification: We have a discussion about the broader impact of our work in Appendix~\ref{app:broader_impact}.
    
\item {\bf Safeguards}
    \item[] Question: Does the paper describe safeguards that have been put in place for responsible release of data or models that have a high risk for misuse (e.g., pretrained language models, image generators, or scraped datasets)?
    \item[] Answer: \answerNA{} 
    \item[] Justification: \answerNA{}

\item {\bf Licenses for existing assets}
    \item[] Question: Are the creators or original owners of assets (e.g., code, data, models), used in the paper, properly credited and are the license and terms of use explicitly mentioned and properly respected?
    \item[] Answer: \answerYes{} 
    \item[] Justification: We cited the appropriate references for each dataset and, when a license is available, we added that information to the citation. We also made it clear that we downloaded all datasets through the torchvision \citep{torchvision2016} python package.

\item {\bf New Assets}
    \item[] Question: Are new assets introduced in the paper well documented and is the documentation provided alongside the assets?
    \item[] Answer: \answerNA{} 
    \item[] Justification: \answerNA{}

\item {\bf Crowdsourcing and Research with Human Subjects}
    \item[] Question: For crowdsourcing experiments and research with human subjects, does the paper include the full text of instructions given to participants and screenshots, if applicable, as well as details about compensation (if any)? 
    \item[] Answer: \answerNA{} 
    \item[] Justification: \answerNA{}

\item {\bf Institutional Review Board (IRB) Approvals or Equivalent for Research with Human Subjects}
    \item[] Question: Does the paper describe potential risks incurred by study participants, whether such risks were disclosed to the subjects, and whether Institutional Review Board (IRB) approvals (or an equivalent approval/review based on the requirements of your country or institution) were obtained?
    \item[] Answer: \answerNA{} 
    \item[] Justification: \answerNA{}

\end{enumerate}